\documentclass{article}
\usepackage[utf8]{inputenc}
\usepackage{authblk}
\usepackage{setspace}
\usepackage[margin=1.25in]{geometry}
\usepackage{graphicx}
\graphicspath{ {./figures/} }
\usepackage{subcaption}
\usepackage{amsmath,amssymb,bm}
\usepackage{lineno}
\usepackage{algorithm}
\usepackage{longtable}
\usepackage{booktabs}
\usepackage{array}
\usepackage{multicol}
\usepackage{commath}
\usepackage{xcolor}

\usepackage[style=ieee, 
citestyle=numeric-comp,
sorting=none]{biblatex}
\title{Bayesian CART models for insurance claims frequency }
\author[1]{Yaojun Zhang 
\thanks{mmyz@leeds.ac.uk}}
\author[1]{Lanpeng Ji 
\thanks{l.ji@leeds.ac.uk}}
\author[1]{Georgios Aivaliotis 
\thanks{G.Aivaliotis@leeds.ac.uk}}
\author[1]{Charles Taylor 
\thanks{c.c.taylor@leeds.ac.uk}}
\affil[1]{\footnotesize Department of Statistics, University of Leeds}

\date{}

\newtheorem{remark}{Remark}

\newcommand{\COM}[1]{}

\newcommand{\vk}[1]{\bm{#1}}
\def\mT{\mathcal{T}}
\def\wkt{{\widehat{\kappa}_t}}
\newcommand{\ga}[1]{\textcolor{black}{#1}}
\newcommand{\yz}[1]{\textcolor{black}{#1}}
\newcommand{\ji}[1]{\textcolor{black}{#1}}
\newcommand{\lj}[1]{\textcolor{black}{#1}}
\newcommand{\yao}[1]{\textcolor{black}{#1}}
\newcommand{\lji}[1]{\textcolor{black}{#1}}
\newcommand{\llj}[1]{\textcolor{black}{#1}}

\allowdisplaybreaks[4]

\begin{document}

\maketitle

\begin{abstract}
\noindent 

The accuracy and interpretability of a (non-life) insurance pricing model are essential qualities to ensure fair and transparent premiums for policy-holders, that reflect their risk. In recent years,  classification and regression trees (CARTs) and their ensembles have gained popularity in the actuarial literature, since they offer good prediction performance and are relatively easy to interpret. In this paper, we introduce Bayesian CART models for insurance pricing, with a particular focus on claims frequency modelling. 
In addition to the common Poisson and negative binomial (NB) distributions used for claims frequency, we implement Bayesian CART for the zero-inflated Poisson (ZIP) distribution to address the difficulty arising from the imbalanced insurance claims data. To this end, we introduce a general MCMC algorithm using data augmentation methods for posterior tree exploration. 
We also introduce the deviance information criterion (DIC) for tree model selection. The proposed models are able to identify trees which can better classify the policy-holders into risk groups. Simulations and real insurance data will be used to illustrate the applicability of these models.
\medskip

\end{abstract}


\section{Introduction}

An insurance policy refers to an agreement between an insurance company (the insurer) and a policy-holder (the insured), in which the insurer promises to charge the insured a certain fee for some unpredictable losses of the customer within a period of time, usually one year. The charged fee is called a {\it premium} which includes a {\it pure premium} and other loadings such as operational costs. 
For each policy, the pure premium is determined by multiple explanatory variables (such as characteristics of the policy-holders, the insured objects, the geographical region, etc.), also called  \textit{risk factors} \cite{ohlsson2010non}.
The premium charged reflects the customer's degree of risk; a higher premium suggests a \ji{potential higher risk}, 
and vice versa.  Therefore, it is necessary to use risk factors to classify policy-holders with similar risk profiles into the same tariff class. The insureds in the same group, all having similar risk characteristics, will pay the same reasonable premium. The process of constructing these tariff classes is also known as {\it risk classification}; see, e.g., \cite{denuit2007actuarial, henckaerts2018data}.
In the basic formula of non-life insurance pricing, the pure premium is obtained by multiplying the expected claims frequency with the conditional expectation of severity, assuming  independence between frequency and severity; see, e.g., \cite{henckaerts2021boosting}. Hence, modelling the claims frequency represents an essential first step in non-life insurance pricing. In this paper, we propose efficacious approaches (namely, Bayesian CARTs or BCART models) to analyze imbalanced insurance claims frequency data. 

Due to its flexibility in modelling a large number of distributions in the exponential family,  generalized linear models (GLMs), developed in \cite{nelder1972generalized}, have been the industry-standard predictive models for insurance pricing \cite{denuit2007actuarial, Wuthrich2022}. 
Explanatory variables enter a GLM through a linear predictor, leading to interpretable effects of the risk factors on the response. Extensions of GLMs to generalized additive models (GAMs) to capture the nonlinear effects of risk factors sometimes offer more flexible models.
However, both GLMs and GAMs often fail to identify the complex interactions among  risk factors. Another popular classical \ji{method based on Bayesian statistics, the credibility method,} was introduced to deal with multi-level factors and lack of data issues; see, e.g., \cite{ohlsson2010non, buhlmann2005course}. Because of the limitations of these classical statistical methods and equipped with continually developing technologies, further research has recently turned to machine learning techniques.  Several machine learning methods such as neural networks, regression trees, bagging techniques, random forests and boosting machines have been introduced in the context of insurance by adopting actuarial
loss distributions in these models to capture the characteristics of insurance claims. We refer to  \cite{blier2020machine} for a recent literature review on this topic and \cite{denuit2019effective, WuthrichMerz2022b,WuthrichBuser2022} for more detailed discussion.

Insurance pricing models are heavily regulated and they must meet specific requirements before being deployed in practice, which posts some challenges for machine learning methods; see \cite{henckaerts2021boosting}. Therein, it is stressed  that pricing models must be transparent and easy to communicate to all the stakeholders and that the insurer has the social role of creating solidarity among the policy-holders so that the use of machine learning for pricing should in no way lead to an extreme penalization of risk or discrimination. The latter has also been noted recently in, e.g., \cite{denuit2021autocalibration,wuthrich2020bias} where it is claimed that prediction accuracy on an individual level should not be the ultimate goal in insurance pricing; \ji{one also needs to ensure} the balance property. Bearing these points in mind, researchers have concluded that tree-based models are good candidates for insurance pricing \cite{henckaerts2021boosting, quan2019insurance, hu2022imbalanced,meng2022actuarial,lindholm2022local}. More precisely,
 the use of  
 \ji{CART}, first introduced in \cite{breiman1984classification},
 partitions a portfolio of policy-holders into smaller groups of homogeneous risk profiles based on some risk factors in which a constant prediction  is used for each sub-group. This results in a highly transparent model and automatically induces solidarity among the policy-holders in a sub-group. 
Although a large number of scholars have carried out empirical and theoretical studies on the effectiveness of CART, limitations of the forward-search recursive partitioning method used in CART have been identified. In particular, the predictive performance tends to be low, and it is known to be unstable: small variations in the training set can result in \yz{greatly} different trees and different predictions for the same test examples. Due to these limitations, more complex tree-based models that combine multiple trees in an ensemble have been popular in insurance prediction \ji{and pricing}, but these ensemble techniques usually introduce additional difficulties in model transparency. In this paper and the sequel, we propose BCART models for insurance claims prediction. Instead of making an ensemble of trees, we look for one \yz{good}  tree, which can improve the prediction ability whilst 
\ji{ensuring} model transparency, by adopting a Bayesian approach applied to CART. 

\begin{figure}[htbp]
	\centering	\includegraphics[height=8.5cm,width=8.5cm]{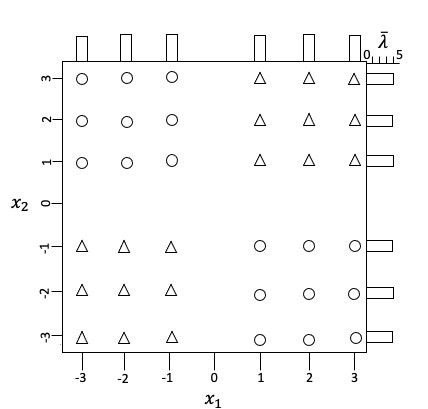}
	\caption{Covariate partition for a Poisson-distributed simulation. \yao{Two covariates \lji{$x_1, x_2$ follow uniform distribution, i.e., $x_1, x_2 \sim U\{-3,-2,-1,1,2,3\}$}. The response variable, which is simulated for the points, has Poisson intensity  equal to  1 (circles) and 7 (triangles). \yao{Each bar represents the average value \lji{($\approx 4$)} of Poisson intensity in that row/column \lji{of data}.}}}
 \label{Fig:S1_data}
\end{figure}

BCART models were first introduced by Chipman et al. \cite{chipman1998bayesian} and Denison et al. \cite{denison1998bayesian}, independently. 
The method has two basic components, prior specification (for the tree and its terminal node parameters) and a stochastic search. The method is to obtain a posterior distribution given the prior, thus leading the stochastic search towards more promising tree models. 
Compared with the tree that CART generates by a greedy forward-search recursive partitioning method, the BCART model generates a 
\ji{much better} tree by an effective Bayesian-motivated stochastic search algorithm. 
This has been justified by simulation examples (with Gaussian-distributed data) in the aforementioned papers. Here, we show another simulation example with Poisson-distributed data to illustrate the effectiveness of BCART. Specifically, we simulate 5,000 Poisson-distributed observations where the Poisson intensity depends on two explanatory variables (or covariates) $x_1$ and $ x_2$ as illustrated in Figure \ref{Fig:S1_data}.
(See also Subsection \ref{subsec:S1} for  \ji{a slightly more general} simulation example.) It is clear from the figure that the optimal partition of the covariate space consists of four regions where the data in each region should follow a homogeneous Poisson distribution. \lji{Note that the ``standard'' CART will not be able to find the correct partition of the data as the Poisson intensities are almost uniform for both marginal distributions 
(see Figure \ref{Fig:S1_data}) 
and no matter how the first split is chosen, it is difficult to distinguish different Poisson intensities on the resulting subsets.}
In contrast, the proposed Poisson BCART can retrieve the optimal tree structure since it has the ability to explore the tree space in a global way (for example, it can modify previously chosen splits).

Since BCART models and their ensemble version -- the Bayesian Additive Regression Trees (BART) models -- generally outperform  other machine learning models, \yz{they have} been extensively studied in the literature; see, e.g., \cite{linero2017review, chipman2010bart, prado2021accounting,  murray2021log, hill2020bayesian} and references therein. In particular, their excellent empirical performance has also motivated works on their theoretical foundations; see \cite{lineroYang2018bayesian, rockova2020posterior}.
However, in most of these studies, the focus has been on Gaussian-distributed data, with some exceptions such as \cite{murray2021log, linero2020semiparametric}. It turns out that a data augmentation approach is needed when dealing with general non-Gaussian data. The existing algorithms do not seem to be directly applicable to insurance data for prediction and pricing. To cover this gap,  as a first step we propose  BCART models for claims frequency  taking account special features of insurance data such as the  high number of zeros and involvement of exposures. We refer to \cite{lee2020delta, lee2021addressing} for a review of claims frequency modelling which also includes some nice \ji{analyses} on exposures. 



The main contributions of this paper are as follows:
\begin{itemize}
    \item We give a general MCMC algorithm for the BCART models applied to any distributed data, where a data augmentation may be needed. In doing so, we follow some ideas in \cite{meng1999seeking, van2001art}.
    \item We introduce a novel model selection method for BCART models based on the deviance information criterion (DIC). 
    Note that DIC was introduced in \cite{spiegelhalter2002bayesian} which appeared a few  years after the introduction of BCART \cite{chipman1998bayesian}. The effectiveness of this approach \ji{is} illustrated by several designed simulation examples and real insurance data. 
    \item We implement the BCART for Poisson, 
 NB and 
    ZIP distributions which are not currently available in any existing R packages. 
    In particular, we introduce two different ways of incorporating exposure in the \ji{NB} and \ji{ZIP} models, following the lines of study in \cite{lee2020delta, lee2021addressing}. The simulation examples and real insurance data analysis show the \ji{applicability} of these proposed BCART models. 

    \item To date, Bayesian tree-based models have not attracted enough attention \ji{compared to} other machine learning methods in the actuarial community. This first step of applying BCART for claims frequency 
\yz{modelling} will open the door for more sophisticated \ji{tree-based} models to meet the needs of the insurance industry. 
\end{itemize}

\textbf{Outline of the rest of the paper}: In Section 2, we review the BCART framework which includes an extension with data augmentation and a model selection method using DIC. Section 3 introduces the notation for insurance claims frequency data and five BCART models including \ji{a} Poisson model, two \ji{NB} models and two ZIP models. In Section 4, we discuss the applicability of the proposed BCART models using three simulation examples and a real insurance claims dataset. Section 5 concludes the paper.

\section{Bayesian CART} \label{Sec_BCART}

We shall briefly review the BCART framework of the seminal paper \cite{chipman1998bayesian}.
 We begin with the general structure of a CART model. Consider a data set $(\vk{X},\vk{y})=\big((\vk{x}_1, y_1),(\vk{x}_2, y_2), \ldots, (\vk{x}_n, y_n)\big)^\top$ with $n$ observations. For the $i$-th observation, $\vk{x}_i=(x_{i1}, x_{i2},\ldots,x_{ip})$ is a vector of $p$ explanatory variables (or covariates) sampled from a space \ji{$\mathcal{X}$,} 
 while $y_i$ is a response variable sampled from a space $\mathcal{Y}$. For our purpose of claims frequency modelling, $\mathcal{Y}$ will be a set of non-negative integers.


A CART has two main components: a binary tree $\mathcal{T}$ with $b$ terminal nodes which induces a partition of the covariate space $\mathcal{X}$, denoted by $\left\{\mathcal{A}_{1}, \ldots, \mathcal{A}_{b}\right\}$, and a parameter \ji{$\vk{\theta}=\left(\vk{\theta}_{1}, \vk{\theta}_{2}, \ldots, \vk{\theta}_{b}\right)$}  which associates the parameter value $\vk{\theta}_t$ with the $t$-th terminal node. Note that here we do not specify the dimension \ji{and range} of the parameter $\vk{\theta}_t$ which should be clear in the considered context below. 
If $\boldsymbol{x_{i}}$ is located in 
the $ t$-th terminal node (i.e., $\vk{x}_i\in \mathcal{A}_t$), then $y_{i}$ 
has a distribution $f\left(y_{i}\mid\vk{\theta}_t\right)$, where $f$ represents a parametric family indexed by $\vk{\theta}_t$. 

By associating observations with the $b$ terminal nodes in the tree $\mathcal{T}$, we can represent the data set as 
$$
(\vk{X},\boldsymbol{y}) =\big((\vk{X}_1,\boldsymbol{y}_{1}),(\vk{X}_2,\boldsymbol{y}_{2}) \ldots, (\vk{X}_b,\boldsymbol{y}_{b})\big)^\top,
$$
where $\boldsymbol{y_{t}}=\left(y_{t 1}, \ldots y_{t n_{t}}\right)^\top$ with $n_{t}$ denoting the number of observations and $y_{tj}$ denoting the $j$-th observation in the $t$-th terminal node, and $\boldsymbol{X}_t$ is an analogously defined $n_t\times p$ design matrix. 
We shall make the typical assumption that 
conditionally on $(\vk{\theta}, \mathcal{T})$, response variables within a terminal node are independent and identically distributed (IID), and they are also independent across terminal nodes. 
\yao{The CART model likelihood in this case will take the form}
\begin{equation}\label{17}
p(\boldsymbol{y} \mid \boldsymbol{X}, \vk{\theta}, \mathcal{T})=\prod_{t=1}^{b} f\left(\boldsymbol{y_{t}} \mid \vk{\theta}_t\right)=\prod_{t=1}^{b} \prod_{i=1}^{n_{t}} f\left(y_{ti} \mid \vk{\theta}_t\right).
\end{equation}
\lji{It is worth noting that instead of the IID assumption within the terminal nodes more general models can be considered, see, e.g., \cite{chipman2002bayesian, chipman2003bayesian} and the references therein. 
}

\yao{Given that $(\vk{\theta}, \mathcal{T})$ determines a CART model, a Bayesian analysis of the problem is conducted by specifying a prior distribution $p(\vk{\theta}, \mathcal{T})$,} \lji{and inference about $\vk{\theta}$ and $\mathcal{T}$ will be based on the joint posterior $p(\vk{\theta}, \mathcal{T}| \vk{y})$ using a suitable MCMC algorithm. 
} Since $\vk{\theta}$ indexes the parametric model whose dimension depends on the number of terminal nodes of the tree, it is usually convenient to 
apply the relationship
\begin{equation}
p(\vk{\theta}, \mathcal{T})=p(\vk{\theta}\mid\mathcal{T}) p(\mathcal{T})
\end{equation}
and specify the tree prior distribution $p(\mathcal{T})$ and the terminal node parameter prior distribution $p(\vk{\theta}\mid\mathcal{T})$, respectively. 
\yao{This strategy, introduced by \cite{george1998}, offers several advantages for Bayesian model selection as outlined in \cite{chipman1998bayesian}.  
}

\subsection{Specification of tree prior $p(\mathcal{T})$} \label{Sec:priorT}

The prior for $\mathcal{T}$ has two components: a tree topology and a decision rule for each of the internal/branch nodes.
\lji{We shall adopt the branching process prior for the topology of $\mathcal{T}$ proposed by Chipman et al. \cite{chipman1998bayesian}.  Due to its computational effectiveness using Metropolis-Hastings \llj{(MH)} search algorithms, this prior specification has been the most popular in the literature.}
A draw from this prior is obtained by generating, for each node  at depth $d$ (with $d=0$ for the root node), two child nodes with probability
\begin{equation}\label{eq:qd}
p(d)=\gamma\left(1+d\right)^{-\rho}, 
\end{equation}
where $\gamma>0, \rho \geq 0$ are parameters 
controlling the \yz{structure} and size of the tree. 
This process iterates for $d=0,1, \ldots,$ until we reach a depth  at which all the nodes 
cease growing.
Note that $p(d)$ is not a probability mass function, but instead is the probability of a given node at depth $d$ being converted to a branch node. A sufficient condition for the termination of this branching process is that $\rho>0$, \yz{and} the case $\rho=0$ corresponds to the Galton-Watson process, see, e.g., \cite{athreya2004branching}. We refer to \cite{linero2018bayesian} for further theoretical discussion of this prior. \lji{Clearly, $\gamma$ controls the overall rate of branching at a node, and the larger $\rho$ becomes, the less likely that deeper nodes will branch, resulting in relatively smaller trees.  In \cite{chipman1998bayesian}, some simulations about the number of terminal nodes associated with the values of the pair $(\gamma, \rho)$ are carried out, which have been used as a guidance when choosing these parameters to generate trees with a certain number of terminal nodes.}

After the tree topology is generated, each internal node is associated with a decision rule 
of the form $x_{l} < c_{l}$ or $x_l\in C_{l}$ according to whether $x_{l}$ is a continuous or a categorical explanatory variable, where  $x_l$ is selected independently and uniformly among the available explanatory variables for each internal node, and the split value $c_l$ or split category subset $C_l$ are selected uniformly among those available for the selected variable $x_l$. In practice, we only consider the overall set of possible split values to be finite; if the $l$-th variable is continuous, the grid for the variable is either uniformly spaced or given by a collection of observed quantiles of $\{x_{il}, i=1,2,\ldots,n\}$. 
\lji{If the $l$-th variable is categorical, the split category subset $C_l$ is usually
selected uniformly among all possible subsets. However, this approach may not be efficient in the (Bayesian) tree search, particularly when the number of categorical levels of $x_l$ is large. Instead, we shall adopt the same treatment of categorical variables as  in the traditional CART greedy search algorithm.
For example, in the Poisson case this is done as follows: calculate for each available categorical level, say $k$, of $x_l$ in that node the empirical frequency $\bar \lambda_k(x_l)$ and use this empirical frequency $\bar\lambda_k(x_l)$ as a numerical replacement for the categorical level $k$ of $x_l$. A subset $C_l$ will be selected uniformly based on the ordered values $\bar\lambda_k(x_l)$.}


\yao{Certainly,} \lji{ 
the design of tree prior can be more intricate
than the one proposed in \cite{chipman1998bayesian}. There  have been several alternatives discussed in the literature.
In a recent contribution \cite{rockova2020posterior}, the convergence of the posterior distribution with a near-minimax concentration rate is studied, where it is shown that the original proposal given by \eqref{eq:qd} does not decay at a fast enough rate to guarantee the optimal rate of convergence.
Instead, a sufficient condition for optimality is induced by the following probability
$$
p(d)=\gamma^{d}, \qquad \text{for some } 0<\gamma<1/2.
$$
Most recently, it is noted  in \cite{saha2023theory} that  the original proposal \eqref{eq:qd} can still offer better empirical solutions. We believe further theoretical and empirical studies in this direction are still needed.
An alternative to the branching process prior is to  specify a prior directly on the number of leaves and a conditionally-uniform prior on the space of trees. 
In \cite{denison1998bayesian}, a Poisson-distributed prior is used for the number of leaves, and then a uniform prior over valid trees (i.e., trees with no empty bottom leaves) with that number of leaves is imposed. As noticed by \cite{wu2007bayesian}, the uniform prior over valid trees in \cite{denison1998bayesian} tends to produce more unbalanced trees 
than balanced ones. Instead, they propose a pinball prior which can generate balanced or skewed trees by adjusting a hyper-parameter. Furthermore, instead of uniformly selecting 
the split value, 
a normal distribution is used for the split value in their simulation and real data analysis in \cite{wu2007bayesian}. Recently, some other tree priors have also been introduced for the purpose of variable selection (particularly when $p>n$), see, e.g., \cite{bleich2014variable,linero2018bayesian,rockova2020posterior, liu2021variable}.
In \cite{linero2018bayesian}, the author proposes a  
sparsity-inducing Dirichlet prior for 
the splitting proportions of the explanatory variables, resulting in this prior allows the model to perform a fully Bayesian variable selection. Furthermore, in \cite{rockova2020posterior, liu2021variable} a spike-and-tree variant is proposed by injecting one more layer on top of the prior used in \cite{denison1998bayesian}, that is, a prior over the active set of explanatory variables. }

\lji{In our current implementation, we adopt the uniform specification for both variable and split value in each of the internal nodes, which is natural and simple. It is also noted in \cite{chipman1998bayesian} that it would be beneficial to incorporate expert knowledge on the prior specification (i.e., using a non-uniform prior), however, our simulation studies in Section \ref{subsec:S1}  show that using the uniform prior  is able to identify the correct splitting rules even in the presence of noise variables. This seems to be a consequence of the Metropolis-Hastings 
random search steps, which tends to not accept noise splitting variables. We refer to \cite{bleich2014variable} for some relevant discussions with the same conclusion.
}

\subsection{Specification of the terminal node parameter prior $p(\vk{\theta}\mid\mathcal{T})$ } \label{subsec_para_prior}

When choosing 
$p(\boldsymbol{\theta}\mid\mathcal{T})$, it is 
\yz{vital} to realize that 
\yz{employing} priors that allow for analytical simplification can greatly reduce the computational burden of posterior calculation and exploration. This is especially true for the choice of the form $p(\boldsymbol{\theta}\mid\mathcal{T})$ for which it is possible to analytically margin out $\vk{\theta}$ to obtain the integrated likelihood
\begin{eqnarray}
\label{eq:int_lik_1}
p(\boldsymbol{y} \mid \boldsymbol{X}, \mathcal{T})&=&\int p(\boldsymbol{y} \mid \boldsymbol{X}, \vk{\theta}, \mathcal{T}) p(\boldsymbol{\theta}\mid\mathcal{T}) d {\vk{\theta}}= \prod_{t=1}^{b}  \int f\left(\vk y_{t} \mid \vk{\theta}_t\right) p(\vk{\theta}_t) d\vk{\theta}_t\nonumber\\
&=&   \prod_{t=1}^{b} \int \prod_{i=1}^{n_{t}} f\left(y_{ti} \mid \vk{\theta}_t\right) p(\vk{\theta}_t) d\vk{\theta}_t,
\end{eqnarray}
where in the second equality we assume that conditional on the tree $\mathcal{T}$ with $b$ terminal nodes as above, the parameters $\vk{\theta}_t, t=1,2,\ldots,b$, have IID priors $p(\vk{\theta}_t)$, which is a common assumption.
Examples where this integration has \yz{a} closed-form expression can be found in, e.g., \cite{chipman1998bayesian, linero2017review}, particularly for Gaussian-distributed data $\vk{y}$. When no such priors can be found, we have to resort to the technique of data augmentation (see, e.g., \cite{kindo2016multinomial, linero2020semiparametric, murray2021log}) which will  be discussed later.
Combining the integrated likelihood $p(\boldsymbol{y} \mid \boldsymbol{X}, \mathcal{T})$ with tree prior $p(\mathcal{T})$, allows us to calculate the posterior of $\mathcal{T}$ 
\begin{equation}\label{eq:post_T}
p(\mathcal{T} \mid \boldsymbol{X}, \boldsymbol{y}) \propto p(\boldsymbol{y} \mid \boldsymbol{X}, \mathcal{T}) p(\mathcal{T}).
\end{equation}

When using MCMC to conduct Bayesian inference, $\mathcal{T}$ can be updated using an \ji{MH} algorithm with the right-hand side of \eqref{eq:post_T} used to compute the acceptance \yz{ratio}. These MH simulations 
can be used to stochastically search the posterior space over trees to determine the high posterior probability trees from which we can choose \yz{a best} one. 
\yz{The posterior sequence for $\vk{\theta}$ is then obtained using an additional Gibbs sampler.} It is worth noting that by integrating out $\vk\theta$ in \eqref{eq:int_lik_1} we avoid the possible complexities associated with reversible jumps between continuous spaces of varying dimensions \cite{chipman2010bart, green1995reversible}.

\subsection{Stochastic search of posterior trees and parameters}
Starting from the root node, the MCMC algorithm for simulating a Markov chain sequence of pairs $\left(\vk{\theta}^{(1)}, \mathcal{T}^{(1)}\right), \left(\vk{\theta}^{(2)}, \mathcal{T}^{(2)}\right), \ldots,$ using the posterior given in \eqref{eq:post_T}, is given in Algorithm 1.

\begin{algorithm} 
	\caption{One step of the MCMC algorithm for updating \yz{the} BCART parameterized by $(\vk{\theta}, \mathcal{T})$} 
	\hspace*{0.02in} {\bf Input:}
	Data $(\vk{X}, \vk{y})$ and current values $\left(\vk{\theta}^{(m)}, \mathcal{T}^{(m)}\right)$ \\
	\hspace*{0.3in} {\bf 1:}
	Generate a candidate value \(\mathcal{T}^{*}\) with probability distribution \(q\left(\mT^{(m)}, \mT^{*}\right)\)\\
	\hspace*{0.3in} {\bf 2:}
	Set the acceptance ratio $\alpha\left(\mT^{(m)}, \mT^{*}\right)=\min \left\{\frac{q\left(\mT^{*}, \mT^{(m)}\right)}{q\left(\mT^{(m)}, \mT^{*}\right)} \frac{p\left(\vk{y} \mid \vk{X}, \mT^{*}\right)}{p\left(\vk{y} \mid \vk{X}, \mT^{(m)}\right)}
\frac{p\left(\mT^{*}\right)}{p\left(\mT^{(m)}\right)}, 1\right\}$\\
	\hspace*{0.3in} {\bf 3:}
	Update \(\mT^{(m+1)}=\mT^{*}\) with probability  $\alpha\left(\mT^{(m)}, \mT^{*}\right)$, otherwise, set $\mT^{(m+1)}=\mT^{(m)}$\\
	\hspace*{0.3in} {\bf 4:}
	Sample $\vk\theta^{(m+1)} \sim p\left(\vk\theta \mid \mT^{(m+1)},\vk{X}, \vk{y}\right)$\\
 \hspace*{0.02in} {\bf Output:} 
	New values  $\left(\vk{\theta}^{(m+1)}, \mathcal{T}^{(m+1)}\right)$

\label{Alg:1}
\end{algorithm}
In Algorithm \ref{Alg:1}, commonly used proposals (or transitions) for $q(\cdot,\cdot)$ include grow, prune, change and swap (see \cite{chipman1998bayesian}), which are usually selected equal probability (i.e., $1/4$ each). Other proposals have been suggested to improve the mixing of simulated trees, but these are often 
\yz{difficult to put into practice}; see, e.g., \cite{wu2007bayesian, pratola2016efficient}. One of the appealing features of these four proposals is that grow and prune steps are reversible counterparts of one another and both change and swap steps are independently reversible. As noticed in \cite{chipman1998bayesian}, this is very \yz{attractive} for the calculation of 
\ji{$\alpha\left(\mT^{(m)}, \mT^{*}\right)$ in Algorithm \ref{Alg:1}, since there are substantial cancellations in the  ratio} 
(see also \cite{kapelner2013bartmachine} for detailed calculations).  In our implementation, we consider these four proposals detailed as follows:


\begin{itemize}
\item Grow: Randomly select a terminal node. 
Split it into two new child nodes and randomly assign it a decision rule \llj{according to the prior specified in Section \ref{Sec:priorT} until the resulting two child nodes  satisfy a minimum observation requirement. If no such decision rule exists, draw a new terminal node (without replacement) and try again. If no such terminal node exists, stop grow.}

\item Prune: A terminal node is randomly selected. The chosen node and its sibling node are pruned into the direct parent node which then becomes a new terminal node. 
\item Change: 
\llj{apply one of the following two types of change to a selected internal node:}
\begin{itemize}
    \item \lji{Change1: Reassign randomly only the split value/category subset according to the prior specified in Section \ref{Sec:priorT}}. 
    \item \lji{Change2: Reassign randomly both the splitting variable and the corresponding split value/category subset according to the prior specified in Section \ref{Sec:priorT}}.
\end{itemize}
\yao{In each of the above changes, randomly select an internal node with the reassignment selected at random from a set (without replacement) until the updated nodes  satisfy the minimum observation requirement. If no such reassignment exists, draw a new internal node (without replacement) and try again. If no such internal node exists, stop change.}
\item Swap: Randomly pick a parent-child pair which are both internal nodes
and swap their decision rules 
\llj{until the updated nodes  satisfy 
the minimum observation requirement. 
If no such parent-child pair exists, stop swap. }
\end{itemize}

\begin{remark}\label{Rem:alg1}

(a). Note that in step 4 of Algorithm \ref{Alg:1}, sampling of $\vk\theta^{(m+1)}$ is needed only for those nodes that were involved in the proposed move from $\mathcal{T}^{(m)}$ to $\mathcal{T}^*$ and only when this move \yz{was} accepted. 

(b). \lji{In comparison to \cite{chipman1998bayesian}, we apply two types of change moves as discussed in \cite{denison1998bayesian}. The introduction of these two types of change is helpful to improve the mixing of posterior trees, 
\llj{as demonstrated by our simulation study in Section \ref{subsec:S1}. Moreover, it is noted that a swap between a parent-child pair with splits using the same variable is impossible. Considering this in our implementation improves the computational efficiency. 
}}

\end{remark}

\subsection{MCMC algorithm with data augmentation}

In this section, we discuss the case where there is no obvious prior distribution $p(\vk{\theta}_t)$  such that  the integration in \eqref{eq:int_lik_1} is of closed-form, particularly, for non-Gaussian data $\vk y$. In this case, we shall use a data augmentation method in implementing the MCMC algorithm. Some special cases have been discussed in  
\cite{chipman2010bart, kindo2016multinomial, linero2020semiparametric, murray2021log}. 

The term data augmentation 
originated from Tanner and Wong's data augmentation algorithm \cite{tanner1987calculation}. It is introduced purely for computational purposes and a latent variable is required so that the original distribution is the marginal distribution of the augmented one. We refer to \cite{van2001art} for an overview of data augmentation and relevant theory. For our purpose, we augment the data $\vk{y}$ by introducing a latent variable $\vk{z}=(z_1,z_2,\ldots,z_n)$ so that the integration in \eqref{eq:int_lik_2} below is computable for augmented data $(\vk{y}, \vk{z})$. To this end, we shall follow the idea of marginal augmentation  introduced in \cite{meng1999seeking} 
 (see also \cite{van2001art}). In their framework, our parameter $\vk{\theta}$ can be interpreted as a working parameter, and thus the integrated likelihood is given as
\begin{eqnarray}\label{eq:int_lik_2_0}
p(\boldsymbol{y} \mid \boldsymbol{X}, \mathcal{T})=
\int    p(\boldsymbol{y}, \vk{z} \mid \boldsymbol{X}, \mathcal{T})  d\vk{z},
\end{eqnarray}
where 
\begin{eqnarray} \label{eq:int_lik_2}
    p(\boldsymbol{y}, \vk{z} \mid \boldsymbol{X}, \mathcal{T})&=&\int p(\boldsymbol{y}, \vk{z} \mid \boldsymbol{X}, \vk{\theta}, \mathcal{T}) p(\boldsymbol{\theta}\mid\mathcal{T}) d {\vk{\theta}}
    =
    \prod_{t=1}^{b}  \int f\left(\vk y_{t}, \vk{z}_t \mid \vk{\theta}_t\right) p(\vk{\theta}_t) d\vk{\theta}_t \nonumber\\
&=&   \prod_{t=1}^{b} \int \prod_{i=1}^{n_{t}} f\left(y_{ti}, z_{ti} \mid \vk{\theta}_t\right) p(\vk{\theta}_t) d\vk{\theta}_t ,
\end{eqnarray}
with $\vk{z}_t=(z_{t1},z_{t2},\ldots,z_{tn_t})$ defined according to the partition of $\mathcal{X}$ and with obvious independence assumed. 
Following Scheme 3 of \cite{meng1999seeking} (see also Section 3 of \cite{van2001art}), we propose the following 
Algorithm \ref{Alg:2} to simulate a Markov chain sequence of pairs $\left(\vk{\theta}^{(1)}, \mathcal{T}^{(1)}\right), \left(\vk{\theta}^{(2)}, \mathcal{T}^{(2)}\right),\ldots,$ starting from the root node. 

\begin{algorithm}
	\caption{One step of the MCMC algorithm for updating \yz{the} BCART parameterized by $(\vk{\theta}, \mathcal{T})$ using data augmentation}
	\hspace*{0.02in} {\bf Input:}
	Data $(\vk{X}, \vk{y})$ and current values $\left(\vk{\theta}^{(m)}, \mathcal{T}^{(m)},\vk{z}^{(m)}\right)$ \\
	\hspace*{0.3in} {\bf 1:}
	Generate a candidate value \(\mathcal{T}^{*}\) with probability distribution \(q\left(\mT^{(m)}, \mT^{*}\right)\)\\
        \hspace*{0.3in} {\bf 2:}
	Sample $\vk{z}^{(m+1)} \sim p(\vk{z}\mid \vk{X}, \vk{y}, \vk{\theta}^{(m)}, \mathcal{T}^{(m)})$\\
	\hspace*{0.3in} {\bf 3:}
	Set the acceptance ratio $\alpha\left(\mT^{(m)}, \mT^{*}\right)=\min \left\{\frac{q\left(\mT^{*}, \mT^{(m)}\right)}{q\left(\mT^{(m)}, \mT^{*}\right)} \frac{p\left(\vk{y},\vk{z}^{(m+1)} \mid \vk{X}, \mT^{*}\right)}{p\left(\vk{y},\vk{z}^{(m)} \mid \vk{X}, \mT^{(m)}\right)}
\frac{p\left(\mT^{*}\right)}{p\left(\mT^{(m)}\right)}, 1\right\}$\\
	\hspace*{0.3in} {\bf 4:}
	Update \(\mT^{(m+1)}=\mT^{*}\) with probability  $\alpha\left(\mT^{(m)}, \mT^{*}\right)$, otherwise, set $\mT^{(m+1)}=\mT^{(m)}$\\
	\hspace*{0.3in} {\bf 5:}
	Sample $\vk\theta^{(m+1)} \sim p\left(\vk\theta \mid \mT^{(m+1)},\vk{X}, \vk{y}, \vk{z}^{(m+1)}\right)$\\
 \hspace*{0.02in} {\bf Output:} 
	New values  $\left(\vk{\theta}^{(m+1)}, \mathcal{T}^{(m+1)}, \vk{z}^{(m+1)} \right)$
 \label{Alg:2}
\end{algorithm}

 Note that in some cases introducing one latent variable $\vk z$ is 
\yz{insufficient} to obtain a closed-form for the integration in \eqref{eq:int_lik_2}; 
\yz{more latent variables may be required}. In that case, we can easily extend  Algorithm \ref{Alg:2} to include multivariate latent variables and use the Gibbs sampler in step 2. 
\yz{Clearly}, the more latent variables used, the slower the convergence of the Markov chain sequence. As discussed in \cite{van2001art}, it is an ``art'' to search for efficient data augmentation schemes. We discus this point later for the claims frequency models.

\begin{remark}\label{Rem:alg2}
    Similar to Algorithm \ref{Alg:1}, \lj{in step 2 and step 5 of Algorithm \ref{Alg:2} the sampling 
    is needed only for those nodes that were involved in the proposed  move from $\mathcal{T}^{(m)}$ to $\mathcal{T}^*$,  and step 5  is needed only when this move was accepted. 
    }
\end{remark}

\subsection{Posterior tree selection and  prediction} \label{Sec:post_pre}

The MCMC algorithms described in the previous section can be used to search for desirable trees. However, as discussed in \cite{chipman1998bayesian} and illustrated below in our analysis, the algorithms quickly 
converge and then move locally in that region for a long time, which occurs because proposals make local moves over a sharply peaked multimodal posterior.  Instead of making long runs of search to move from one mode to another better one, we follow the idea of \cite{chipman1998bayesian} to repeatedly restart the algorithm. As many trees are visited by each run of the algorithm, we need a method to identify those trees which are of most interest. Moreover, the structure of trees in the convergence regions is mostly determined by the hyper-parameters $\gamma, \rho$ which also need to be chosen appropriately.  In \cite{chipman1998bayesian}, the integrated likelihood $p(\vk{y}\mid \vk{X},\vk{T})$ is used 
as a measure to choose good trees from one run of the algorithm, though other measures, like residual sum of squares, could also be introduced. However, there is no discussion on how the tree prior hyper-parameters $\gamma, \rho$ should be determined \yz{optimally}.  A natural way to deal with this is to use cross-validation which, however, requires repeated model fits and is very computationally expensive. In this paper, we propose to use \lj{DIC} for choosing appropriate $\gamma, \rho$, and thus introduce a three-step approach for selecting an ``optimal'' tree among those visited. To this end, we first give a definition of DIC for a Bayesian \yz{CART}. We refer to \cite{spiegelhalter2002bayesian, celeux2006deviance, gelman2014understanding, spiegelhalter2014deviance} for more detailed discussion of DIC and its extensions. 

Consider the tree $\mathcal{T}$ with $b$ terminal nodes and parameters $\vk{\theta}_t, t=1,2,\ldots,b$, previously defined. We first introduce DIC for each node using the standard definition, the DIC for the tree is then defined as the sum of the DIC of all \ga{terminal} nodes in the tree due to the independence assumption. For node $t$, we call 
\begin{eqnarray} \label{eq:dev_t}
    D(\vk{\theta}_t)=-2\log(f(\vk{y}_t\mid \vk{\theta}_t))=-2 \sum_{i=1}^{n_t} \log(f(y_{ti}\mid \vk{\theta}_t))
\end{eqnarray} 
the {\it deviance}. 

Analogously to Akaike's information criterion (AIC), Spiegelhalter et al. \cite{spiegelhalter2002bayesian} proposed the DIC based on the principle 
DIC$=$``goodness of fit''$+$``complexity'', which is defined as
\begin{eqnarray*}
     \mathrm{DIC}_t= D(\overline{\vk{\theta}_t})+2p_{Dt}, 
\end{eqnarray*}
where $\overline{\vk{\theta}_t}=E_{\text{post}}(\vk{\theta}_t)$ is the posterior mean (with $E_{\text{post}}$ denoting expectation over the posterior distribution of $\vk{\theta}$ given data \yz{$\vk{y}$}), and $p_{Dt}$ is the 
{\it effective number of parameters} given by
\begin{eqnarray}\label{eq:pDt}
    p_{Dt}=\overline{D(\vk{\theta}_t)}-D(\overline{\vk{\theta}_t})&= &- 2 E_{\text{post}}(\log(f(\vk{y}_t\mid \vk{\theta}_t))) +2 \log(f(\vk{y}_t\mid \overline{\vk{\theta}_t})) \nonumber\\
    &=& 2 \sum_{i=1}^{n_t} \left(\log(f(y_{ti}\mid \overline{\vk{\theta}_t}))-  E_{\text{post}}(\log(f(y_{ti}\mid \vk{\theta}_t))) \right). 
\end{eqnarray}
The DIC of the tree $\mathcal{T}$ with $b$ terminal nodes is then defined as 
\begin{eqnarray}\label{eq:DIC}
     \mathrm{DIC}:=\sum_{t=1}^b \mathrm{DIC}_t=D(\overline{\vk{\theta}})+2 p_D,
\end{eqnarray}
where $D(\overline{\vk{\theta}})=\sum_{t=1}^b D(\overline{\vk{\theta}_t})$ and $p_D=\sum_{t=1}^bp_{Dt}$ are the deviance and effective number of parameters of the tree.

Next, we introduce DIC for tree models with data augmentation. Depending on whether the latent variable $\vk{z}$ is treated as a parameter or not, there are three types of likelihoods leading to eight versions of DIC as discussed in \cite{celeux2006deviance}. 
Due to the complexity in implementing any of those eight and motivated by the idea that 
DIC$=$``goodness of fit''$+$``complexity'', we introduce a new DIC for node $t$ in the tree as follows 
\begin{eqnarray}\label{eq:DICt}
     \mathrm{DIC}_t= D(\overline{\vk{\theta}_t})+2q_{Dt}, 
\end{eqnarray}
where $D(\overline{\vk{\theta}_t})$ is the deviance defined through the data $\vk{y}_t$ (as in \eqref{eq:dev_t}) which represents the goodness of fit, 
and $q_{Dt}$ is the  {\it effective number of parameters} defined through the augmented data $(\vk{y}_t, \vk{z}_t)$ as follows
\begin{eqnarray}\label{eq:QDt}
    q_{Dt}
    &=& - 2 E_{\text{post}}(\log(f(\vk{y}_t,\vk{z}_t\mid \vk{\theta}_t))) +2 \log(f(\vk{y}_t,\vk{z}_t\mid \overline{\vk{\theta}_t}))
    \nonumber\\
    &=& 2 \sum_{i=1}^{n_t} \left(\log(f(y_{ti},z_{ti}\mid \overline{\vk{\theta}_t}))-  E_{\text{post}}(\log(f(y_{ti},z_{ti}\mid \vk{\theta}_t))) \right),
\end{eqnarray}
where $\overline{\vk{\theta}_t}=E_{\text{post}}(\vk{\theta}_t)$, and in this case $E_{\text{post}}$ denotes expectation over the posterior distribution of $\vk{\theta}$ given augmented data $(\vk{y}, \vk{z})$.
As we will see below, for the frequency models, $q_{Dt}$ is approximately the dimension of $\vk{\theta}_t$ as the sample size $n_t$ in node $t$  tends to infinity. Similarly, the DIC of tree $\mathcal{T}$ with $b$ terminal nodes is thus defined as 
\begin{eqnarray} \label{eq:DIC_aug}
     \mathrm{DIC}=D(\overline{\vk{\theta}})+2 q_D,
\end{eqnarray}
where $q_D=\sum_{t=1}^b q_{Dt}$.

\begin{remark}
(a). Note that DIC is defined using plug-in prediction densities $f(y_{ti}\mid \overline{\vk{\theta}_t})$ in \eqref{eq:pDt} (similarly  $f(y_{ti}, z_{ti}\mid \overline{\vk{\theta}_t})$ in \eqref{eq:QDt}). More recently, a new criterion called WAIC 
was introduced by Watababe \cite{watanabe2010asymptotic} (see also \cite{gelman2014understanding,spiegelhalter2014deviance}), where in its definition the plug-in prediction density is replaced by the full prediction density $E_{\text{post}}(f(y_{ti}\mid \vk{\theta}_t))$. When the explicit expression is not available, this posterior expectation is usually computed by a Monte Carlo algorithm as $S^{-1}\sum_{k=1}^S f(y_{ti}\mid \vk\theta^k)$, where $\vk\theta^k$ is simulated from the posterior distribution of $\vk{\theta}_t$. \yz{In the following section, we will see that this  posterior expectation can be obtained explicitly for the Poisson model, but not for other  models.}
It turns out that using WAIC  \yz{gives the same  \lj{selected model as  DIC}} in our simulation examples. \ji{Additionally}, since it involves Monte Carlo  \yz{algorithm} and as such could be considerably more computationally expensive, we suggest \yz{using} DIC.

(b). \llj{It is worth noting that if the independence assumption within the terminal nodes is violated (e.g., \cite{chipman2002bayesian, chipman2003bayesian}), the DIC may also be used as a tool for model selection but the formulation would not be of the simple summation form as in \eqref{eq:dev_t}. We refer to \cite{spiegelhalter2002bayesian} for examples and relevant discussions.}

\end{remark}

Now, we are ready to introduce the  three-step approach for selecting an ``optimal'' tree from the MCMC algorithms. Let $m_s<m_e$ be two user input integers which \ji{represent} the belief that the optimal number of \yz{terminal} nodes lies in $[m_s,m_e]$. In practice, \ji{these} can be estimated \yz{first} by using some other methods, e.g., a standard CART model. 
The three-step approach is 
\yz{described} in Table \ref{table_SS}. \yz{In what follows, the tree selected by using the three-step approach will be called an ``optimal'' tree.}

\begin{table}[!t] 

 \centering

 \caption{Three-step approach for ``optimal'' tree selection}
 \begin{tabular}{p{1.5cm}|p{12cm}}  

\toprule   

{\bf Step 1:} & Set a sequence of hyper-parameters $(\gamma_j, \rho_j), j=m_s, \ldots, m_e,$ such that for $(\gamma_j, \rho_j)$, the MCMC algorithm converges to a region of trees which have $j$ terminal nodes.\\
\hline
{\bf Step 2:} &
For each $j$ in Step 1, select the tree with maximum likelihood $p(\vk{y}\mid \vk{X}, \vk{\overline{\vk{\theta}}}, \mathcal{T})$ from the convergence region. \\  
\hline

{\bf Step 3:} & From the trees obtained in Step 2, select the optimal one using DIC.\\
  \bottomrule  
  
\end{tabular}

  \label{table_SS}
\end{table}

\begin{remark}
(a).    The relation between hyper-parameters $(\gamma_j, \rho_j)$ and the distribution of the number of \yz{terminal} nodes of tree has been illustrated in \cite{chipman1998bayesian}. It does not seem hard to set values for $(\gamma_j, \rho_j)$ so that the MCMC algorithm\yz{s} will converge to a region of trees with required $j$ terminal nodes. It is also worth noting that the distribution of the number of \yz{terminal} nodes  is also affected by the data in hand, which can be seen from the calculation of the acceptance ratio in the MCMC algorithms. In our simulations and real data analysis below, we have to select a relatively larger $\rho$ 
in order to achieve our goals. 

(b). \ji{In Step 2, the so-called data likelihood $p(\vk{y}\mid \vk{X}, \vk{\overline{\vk{\theta}}}, \mathcal{T})$, rather than the integrated likelihood $p(\vk{y}\mid \vk{X},\mathcal{T})$, is used, which is due to our interest in the fit of the parametric model to data. The simulations and real data in Section \ref{sec:sim} indicate that these two types of likelihood show a  consistency in the ordering of their values, and thus we suspect there is no big difference using either of them.}
\end{remark}

Suppose $\mathcal{T}$ with $b$ terminal nodes and parameter $\overline{\vk{\theta}}$ is the optimal tree obtained from the above three-step approach. For a given new $\vk{x}$ the predicted  $\hat{y}$ using this tree model is defined as
\begin{equation}\label{eq:yhat}
    \hat{y}\mid \vk{x} = \sum_{t=1}^b E(y\mid \overline{\vk{\theta}_t}) I_{(\vk{x}\in \mathcal{A}_t)},
\end{equation}
where $I_{(\cdot)}$ denotes the indicator function \lj{and $\{\mathcal{A}_t\}_{t=1}^b$ is the partition of $\mathcal{X}$ by $\mathcal{T}$}.

\begin{remark}
    An alternative prediction given $\vk{x}$ can be defined using the full predictive density as 
    \begin{equation}\label{eq:yhat_full}
    \hat{y}\mid \vk{x} = \sum_{t=1}^b E_{\text{post}} (E(y\mid \vk{\theta}_t)) I_{(\vk{x}\in \mathcal{A}_t)}.
    \end{equation}
However, for the frequency models 
\yz{the explicit expression} can be found only for the Poisson case, and for other models the Monte Carlo method is needed to estimate the posterior expectation. Thus, we shall use \eqref{eq:yhat} for simplicity.
\end{remark}

\section{Bayesian CART claims frequency models}\label{Sec:CF}

In this section, we introduce the BCART for insurance claims frequency \ji{by specifying the response distribution in the general framework introduced in Section \ref{Sec_BCART}}. We shall discuss three commonly used distributions in the literature to model the claim numbers, namely, Poisson, \ji{NB} and ZIP distribution\yz{s}; see, e.g., \cite{WuthrichMerz2022b, lee2020delta, lee2021addressing}. To this end, we first introduce the claims data. A claims data set with $n$ policy-holders can be described  by $(\vk{X}, \vk{v}, \vk{N})=\big((\vk{x}_1,v_1, N_1), \ldots, (\vk{x}_n,v_n, N_n)\big)^\top$, where 
$\vk{x}_i=(x_{i 1}, \ldots, x_{i p})\in \mathcal{X}$ consists of rating variables (e.g., area, driver age, car brand in car insurance)\yz{;} $N_{i}$ is the number of claims  reported, and $v_{i}\in(0,1]$ is the exposure in yearly units which is used to quantify how long the policy-holder is exposed to risk. The goal is to explain and predict the claims information $N_i$ based on the rating variables $\vk{x}_i$ and the exposure $v_i$ for each individual policy $i$, which leads 
to the {\it claims frequency}, i.e., the number of claims filed per unit \yz{year} of exposure to risk. We will discuss below how this can be done with BCART models.

\subsection{Poisson model}
Consider a tree $\mathcal{T}$ with $b$ terminal nodes as discussed in Section \ref{Sec_BCART}. In a Poisson model, we assume 
\begin{equation*}
    N_i \mid \vk{x}_i,v_i \ \stackrel{\text { }}{\sim} 
  \ \operatorname{Poi}\left(v_{i}\sum_{t=1}^b\lambda_t I_{(\vk{x}_i\in \mathcal{A}_t)} \right)
\end{equation*}
for the $i$-th observation 
where $\mathcal{A}_t$ is a partition of $\mathcal{X}$. Here we use the standard notation $\lambda_t$ for claims frequency rather than the generic notation $\vk{\theta}_t$ for the  parameter in terminal node $t$. Essentially, we have specified the distribution $f(y_i\mid \vk{\theta}_t)$ for terminal node $t$ (see Section \ref{Sec_BCART}) as
\begin{equation}\label{eq:f_P}
    f_{\text{P}}\left(m\mid \lambda_t\right)= {P}\left(N_{i}=m\mid \lambda_t\right)=\frac{e^{-\lambda_t v_{i}} {(\lambda_t v_{i})}^{m}}{m !}, \ \ m=0,1,2,\ldots,
\end{equation}
for the $i$-th observation such that $\vk{x}_i\in 
\mathcal{A}_t$. Note that, for simplicity, here and hereafter, the exposure $v_i$ and $
\vk{x}_{i}$ will be compressed in some \lj{notation}.  
Based on the discussions \ji{in Section \ref{subsec_para_prior}}, 
we choose the gamma prior for $\lambda_{t}$ \yz{with hyper-parameters $\alpha, \beta>0$, that is,}
\begin{equation}\label{eq:p_lambda}
p\left(\lambda_{t}\right)=\frac{\beta^{\alpha} {\lambda_{t}}^{\alpha-1} e^{-\beta \lambda_{t}}}{\Gamma(\alpha)},
\end{equation}
with $\Gamma(\cdot)$ denoting the gamma function. As in Section \ref{Sec_BCART}, for terminal node $t$ we define the associated data  as 
\yz{$\left(\vk{X}_t, \vk{v}_t,\vk{N}_t)=(({X}_{t1},v_{t1}, N_{t1}), \ldots, ({X}_{tn_t}, v_{tn_t},N_{tn_t})\right)^\top$. } With the above gamma prior, the integrated likelihood for terminal node $t$ can be obtained as
\begin{equation}
\begin{aligned}
\label{pois_integrated}
p_{\text {P}}\left(\vk{N}_{t} \mid \vk{X}_{t},\vk{v}_t \right) &=  \int_0^\infty f_{\text{P}}\left(\vk{N}_{t} \mid \lambda_{t} \right) p(\lambda_{t} ) d \lambda_{t} \\
&= \int_0^\infty \prod_{i=1}^{n_{t}} \frac{e^{-\lambda_{t} v_{ti}} {\left(\lambda_{t} v_{ti}\right)}^{N_{ti}}}{N_{ti} !} \frac{\beta^{\alpha} {\lambda_{t}}^{\alpha-1} e^{-\beta \lambda_{t}}}{\Gamma(\alpha)} d \lambda_{t} \\
&= \frac{\beta^{\alpha}\prod_{i=1}^{n_{t}} v_{ti}^{N_{ti}}}{\Gamma(\alpha)\prod_{i=1}^{n_{t}} N_{ti} !} \int_0^\infty \lambda_{t}^{\sum_{i=1}^{n_{t}} N_{ti}+\alpha-1} e^{-(\sum_{i=1}^{n_{t}}v_{ti}+\beta)\lambda_{t}} d \lambda_{t}\\
&= \frac{\beta^{\alpha}\prod_{i=1}^{n_{t}} v_{ti}^{N_{ti}}}{\Gamma(\alpha)\prod_{i=1}^{n_{t}} N_{ti} !}
\frac{\Gamma(\sum_{i=1}^{n_{t}} N_{ti}+\alpha)}{(\sum_{i=1}^{n_{t}}v_{ti}+\beta)^{\sum_{i=1}^{n_{t}} N_{ti}+\alpha}}.
\end{aligned}
\end{equation}
Clearly, from \eqref{pois_integrated}, we see that the posterior distribution of $\lambda_{t}$, conditional on $\vk{N}_t$, 
is given by
\begin{eqnarray}
\label{eq:ltNt}
\lambda_{t} \mid\vk{N}_t\ \   \sim \ \ \text{Gamma}\left(\sum_{i=1}^{n_{t}} N_{ti}+\alpha,\sum_{i=1}^{n_{t}}v_{ti}+\beta\right).
\end{eqnarray}
The integrated likelihood for the tree $\mathcal{T}$ is thus given by
\begin{equation}\label{eq:pois_whole}
    p_{\text{P}}\left(\vk{N} \mid \vk{X},\vk{v},\mathcal{T} \right)=\prod_{t=1}^{b} p_{\text{P}}\left(\vk{N}_{t} \mid \vk{X}_{t},\vk{v}_t \right).
\end{equation}
Next, we discuss the DIC for this tree, focusing on DIC$_t$ for terminal node $t$. First, we have
\begin{equation}\label{eq:Dlam_t}
 D\left(\lambda_{t}\right)=-2 \sum_{i=1}^{n_t}\log f_\text{P}(N_{ti}\mid \lambda_t)
 =-2 \sum_{i=1}^{n_t}\left(-\lambda_{t} v_{ti}+N_{ti} \log \left(\lambda_{t} v_{ti}\right)-\log \left(N_{ti} !\right)\right),
\end{equation}
and by \eqref{eq:ltNt} we get the posterior mean for $\lambda_t$ as
\begin{eqnarray}\label{eq:pois_mean}
\overline{\lambda}_{t}= E_{\text{post}}(\lambda_t)=\frac{\sum_{i=1}^{n_t} N_{ti}+\alpha}{\sum_{i=1}^{n_t} v_{ti}+\beta}.
\end{eqnarray}
Furthermore, we derive that
\begin{eqnarray}\label{eq:mean_Dlam_t}
\overline{D\left(\lambda_t\right)} &=&E_{\text{post}}\left(D\left(\lambda_t\right)\right)\nonumber \\
&=&2 \sum_{i=1}^{n_t} v_{t i} E_{\text{post}}\left(\lambda_t\right)-2  \sum_{i=1}^{n_t} N_{t i} E_{\text{post}}\left(\log \left(\lambda_t\right)+\log \left(v_{t i}\right)\right)+2  \sum_{i=1}^{n_t} \log \left(N_{t i} !\right) \nonumber\\
&=&2  \left( \frac{\sum_{i=1}^{n_t} N_{t i}+\alpha}{\sum_{i=1}^{n_t} v_{t i}+\beta}\right) \sum_{i=1}^{n_t} v_{t i}  -2 \left(\psi\left( \sum_{i=1}^{n_t} N_{t i}+\alpha\right)-\log \left(\sum_{i=1}^{n_t} v_{t i}+\beta\right)\right)  \sum_{i=1}^{n_t} N_{t i} \nonumber\\
&&\ \ \ \  -2  \sum_{i=1}^{n_t} N_{t i}\log \left(v_{t i}\right)+2  \sum_{i=1}^{n_t} \log \left(N_{t i} !\right),
\end{eqnarray}
where we have used the fact that 
$$
E_{\text{post}} \left(\log \left(\lambda_t\right) \right)=\psi\left( \sum_{i=1}^{n_t} N_{t i}+\alpha\right)-\log \left(\sum_{i=1}^{n_t} v_{t i}+\beta\right),
$$ with $\psi(x)= \Gamma'(x)/\Gamma(x)$ being the digamma function.
Using \eqref{eq:Dlam_t}--\eqref{eq:mean_Dlam_t}, we obtain the effective number of parameter\yz{s} for \yz{terminal} node $t$ as
$$
\begin{aligned}
p_{D{t}}&= \overline{D(\lambda_{t})}-D(\overline{\lambda_{t}})\\
&=2
\left(\log \left(\sum_{i=1}^{n_t} N_{ti}+\alpha\right)-\psi\left(\sum_{i=1}^{n_t} N_{ti}+\alpha\right)\right)  \sum_{i=1}^{n_t} N_{ti},
\end{aligned}
$$
and 
$$
\begin{aligned}
\text{DIC}_t &=D\left(\overline{\lambda_t}\right)+2 p_{Dt} \\
&=2 \left(\frac{\sum_{i=1}^{n_t} N_{t i}+\alpha}{\sum_{i=1}^{n_t} v_{t i}+\beta}\right) \sum_{i=1}^{n_t} v_{t i} -2 \sum_{i=1}^{n_t} N_{t i}\left(\log \left(\frac{\sum_{i=1}^{n_t} N_{t i}+\alpha}{\sum_{i=1}^{n_t} v_{t i}+\beta}\right)+\log \left(v_{t i}\right)\right) +2 \sum_{i=1}^{n_t} \log \left(N_{t i} !\right) \\
&\qquad +4 \left(\log \left(\sum_{i=1}^{n_t} N_{t i}+\alpha\right)-\psi\left(\sum_{i=1}^{n_t} N_{t i}+\alpha\right)\right)\ji{\sum_{i=1}^{n_t} N_{t i}}.
\end{aligned}
$$
Then the DIC of tree $\mathcal{T}$ is \yz{obtained using \eqref{eq:DIC}}. 

\begin{remark}
    Since $\psi(x)= \log(x) - \frac{1}{2x} (1+o(x)), \yz{as} \ x \to\infty$, we immediately see that $p_{Dt}\to1$ as  \yz{$n_t \to\infty$}. This explains the name of effective number of parameters in the Bayesian framework, as 1 is the number of parameters in the \yz{terminal} node $t$ \yz{for Poisson model} if a flat prior \ji{is} assumed for $\lambda_t$.
\end{remark}

With the above \eqref{eq:ltNt}--\eqref{eq:pois_whole} and \yz{DIC obtained}
, we can  use the three-step approach proposed in Section \ref{Sec:post_pre} to search for an optimal tree, where  \eqref{eq:ltNt} and \eqref{eq:pois_whole} should be used in step 4 and step 2, respectively, 
in Algorithm 1. Given an optimal tree, the estimated claims frequency $\overline{\lambda_t}$ in terminal node $t$ can be given by the posterior mean in \eqref{eq:pois_mean}, using \eqref{eq:yhat}. 
It is worth noting that we can obtain the same estimate 
by using \eqref{eq:yhat_full} instead.

\subsection{Negative binomial models}\label{NB} 
The \ji{NB} distribution, a member of mixed Poisson \yz{family}, offers an effective way to handle over-dispersed insurance \yz{claims} frequency data \ji{where excessive zeros are common}. 

Consider the tree $\mathcal{T}$ with $b$ terminal nodes as before. In the \ji{NB} model, we assume that
$N_{ti}\mid X_{ti}, v_{ti} $ follows a \ji{NB} distribution for all terminal node\yz{s}, $t=1,\ldots,b$. There are different ways to parameterize the \ji{NB} distribution, particularly with the exposure, see, e.g., \cite{lee2020delta, WuthrichMerz2022b}.  We shall discuss two models below.

\subsubsection{Negative binomial model 1 (NB1)}
We first adopt the most common parameterization of the \ji{NB} distribution, see, e.g., \cite{murray2021log}. That is, for terminal node $t$, 
\begin{eqnarray}\label{eq:NB_dens}
    f_{\text{NB1}}(m\mid\kappa_t,\lambda_t) &=&P(N_{ti}=m\mid\kappa_t,\lambda_t)\nonumber\\
&=&\frac{\Gamma(m+\kappa_t)}{\Gamma(\kappa_t) m!} \left(\frac{\kappa_t}{\kappa_t+\lambda_t v_{ti}}\right)^{\kappa_t} \left(\frac{\lambda_t v_{ti}}{\kappa_t+\lambda_t v_{ti}}\right)^{m},\ \ji{m=0,1,\ldots,}
\end{eqnarray}
\ji{where $\kappa_t,\lambda_t>0.$}
It is easy to show that the mean and variance of $N_{ti}$ are given by
\begin{equation}\label{eq:N_mean}
    E(N_{ti}\mid \kappa_t,\lambda_t) =  \lambda_t v_{ti}, \quad \quad
    \text{Var}(N_{ti}\mid \kappa_t,\lambda_t)=  \lambda_t  v_{ti} \left(1+\frac{ \lambda_t v_{ti}}{\kappa_t}\right).
\end{equation}
 The degree of over-dispersion in relation to the Poisson is controlled by the additional parameter $\kappa_t$ in the \ji{NB} model\yz{, which} converges to the Poisson model as
$\kappa_t \rightarrow \infty$. 

In \ji{NB} regression, the lack of simple and efficient algorithms for posterior computation has seriously limited routine applications of Bayesian approaches. Recent studies make Bayesian approaches appealing by introducing data augmentation \yz{techniques}; see, e.g., \cite{zhou2012lognormal, murray2021log}. In order to save on total computational time of the algorithm and avoid the difficulty \ji{of} finding an appropriate prior for $\kappa_t$ with corresponding data augmentation, we shall treat the parameter $\kappa_t$ as known in the Bayesian framework which can be estimated upfront by using, e.g., the moment matching method. However, in line with the Poisson model, we shall treat $\lambda_t$ as uncertain and use a gamma prior with  corresponding data augmentation. 
Based on the formulas given in \eqref{eq:N_mean}, we can estimate the parameter $\kappa_t$, using the moment matching method, see, e.g., Chapter 2 of \cite{Wuthrich2022} as follows
\begin{equation}\label{eq:NB1_kapp}
\widehat{\kappa}_t = \frac{\widehat{\lambda}_t^2 }{\widehat{V}_t^2 - \widehat{\lambda}_t} \frac{1}{n_t-1} \left( \sum_{i=1}^{n_t} v_{ti}- \frac{\sum_{i=1}^{n_t} v_{ti}^2}{\sum_{i=1}^{n_t} v_{ti}} \right),
\end{equation}
where
\begin{equation} \label{eq:Vlambda}
   \widehat{V}_t^2 =\frac{1}{n_t-1} \sum_{i=1}^{n_t}v_{ti} \left(\frac{N_{ti}}{v_{ti}}-\widehat{\lambda}_t \right)^2,\ \ \ 
\widehat{\lambda}_t =\frac{\sum_{i=1}^{n_t} N_{ti}}{\sum_{i=1}^{n_t} v_{ti}}. 
\end{equation}

Next, introducing a latent variable $\vk{\xi}_t=(\xi_{t1}, \xi_{t2},\ldots, \xi_{tn_t}) \in (0,\infty)^{n_t}$, we can define a data augmented likelihood for the $i$-th data \ji{instance in terminal} node $t$ as
\begin{equation}
\label{nb_augment}
f_{\text{NB1}}\left(N_{ti}, \xi_{ti} \mid \widehat{\kappa}_t,\lambda_{t}\right)=\left(\lambda_{t} v_{ti}\right)^{N_{ti}} e^{-\xi_{ti} \lambda_{t}v_{ti}}  \frac{\wkt^{\wkt} \xi_{ti}^{\wkt+N_{ti}-1} e^{-\xi_{ti} \wkt}}{\Gamma(\wkt) N_{ti} !}.
\end{equation}
It is easily checked that 
integrating over $\xi_{ti}\in (0,\infty)$ in \eqref{nb_augment} yields the marginal distribution \eqref{eq:NB_dens}.  Further, we see that $\xi_{ti}$, given data $N_{ti}$ and parameters, is gamma distributed, i.e.,
\begin{equation}\label{eq:xi_post}
    \xi_{ti}\mid N_{ti}, \widehat{\kappa}_t,\lambda_{t}\  \sim \  \text{Gamma}\left(\wkt+N_{ti}, \wkt+\lambda_t v_{ti} \right).
\end{equation}

Given the data augmented likelihood in \eqref{nb_augment}, the estimated parameter $\wkt$ \ji{using \eqref{eq:NB1_kapp}}, and a conjugate gamma prior for $\lambda_{t}$ 
with   hyper-parameters   $\alpha, \beta>0$ (cf. \eqref{eq:p_lambda}),
we can derive the integrated augmented likelihood for the \ji{terminal} node $t$ as follows
\begin{equation}
\begin{aligned}
\label{nb_integrated}
p_{\text{NB1}}\left(\vk{N}_{t}, \vk{\xi}_t\mid \vk{X}_{t}, \vk{v}_t, \wkt\right) 
&=\int_0^\infty f_{\text{NB1}}\left(\vk{N}_t, \vk{\xi}_t \mid \widehat{\kappa}_t,\lambda_{t}  \right) p(\lambda_{t} ) d \lambda_{t} \\
&=\int_0^\infty \prod_{i=1}^{n_{t}}\left[\left(\lambda_{t} v_{ti}\right)^{N_{ti}} e^{-\xi_{ti} \lambda_{t}v_{ti}}  \frac{\wkt^{\wkt} \xi_{ti}^{\wkt+N_{ti}-1} e^{-\xi_{ti} \wkt}}{\Gamma(\wkt) N_{ti} !}\right]\frac{\beta^{\alpha} {\lambda_{t}}^{\alpha-1} e^{-\beta \lambda_{t}}}{\Gamma(\alpha)} d \lambda_{t} \\
&=\frac{\wkt^\wkt \beta^\alpha}{\Gamma(\wkt) \Gamma(\alpha)} \prod_{i=1}^{n_{t}}  \left[\frac{v_{ti}^{N_{ti}}}{N_{ti} !} \xi_{ti}^{\wkt+N_{ti}-1} e^{-\xi_{ti} \wkt}\right]
\frac{\Gamma(\sum_{i=1}^{n_{t}} N_{ti}+\alpha)}{(\sum_{i=1}^{n_{t}}\xi_{ti}v_{ti}+\beta)^{\sum_{i=1}^{n_{t}} N_{ti}+\alpha}}.
\end{aligned}
\end{equation}
Moreover, from the above we see that the posterior distribution of $\lambda_{t}$ given the augmented data $(\vk{N}_t, \vk{\xi}_t)$, 
is given by 
$$
\lambda_{t} \mid \vk{N}_t, \vk{\xi}_t
\ \sim \ \text{Gamma}\left(\sum_{i=1}^{n_{t}} N_{ti}+\alpha,\sum_{i=1}^{n_{t}}\xi_{ti}v_{ti}+\beta\right).
$$
The integrated augmented likelihood for the tree $\mathcal{T}$ is thus given by
\begin{equation}\label{eq:NB_whole}
    p_{\text{NB1}}\left(\vk{N}, \vk{\xi} \mid\vk{X},\vk{v},\widehat{\vk{\kappa}}, \mathcal{T} \right)=\prod_{t=1}^{b} p_{\text{NB1}}\left(\vk{N}_{t}, \vk{\xi}_t\mid \vk{X}_{t}, \vk{v}_t, \wkt \right).
\end{equation}
Now, we discuss the DIC for this tree. Since we only consider uncertainty for $\vk{\lambda}$ but not for $\vk{\kappa}$, the DIC defined in \eqref{eq:DIC_aug} cannot be adopted directly. Thus, using the idea that 
DIC$=$``goodness of fit''$+$``complexity'', we can introduce a new DIC$_t$ for  terminal node $t$ as follows
\begin{eqnarray*}
     \mathrm{DIC}_t= D(\overline{\lambda_t})+2r_{Dt}.
\end{eqnarray*}
Here, the goodness of fit is given by 
$$
D(\overline{\lambda_t})=-2 \sum_{i=1}^{n_t} \log f_{\text{NB1}}(N_{ti}\mid \wkt, \overline{\lambda_t}),
$$
and the effective number of parameters $r_{Dt}$ is given by 
\begin{eqnarray}\label{eq:rDt_NB1}
    r_{Dt}
    =1+ 2 \sum_{i=1}^{n_t} \left(\log(f_{\text{NB1}}(N_{ti},\xi_{ti}\mid \wkt, \overline{\lambda_t}))-  E_{\text{post}}(\log(f_{\text{NB1}}(N_{ti},\xi_{ti}\mid \wkt,\lambda_t))) \right),
\end{eqnarray}
where 1 represents the number for $\kappa_t$ and the second part is for  $\lambda_t$, 
$$\overline{\lambda_t}= \frac{\sum_{i=1}^{n_t} N_{ti}+\alpha}{\sum_{i=1}^{n_t} \xi_{ti}v_{ti}+\beta},$$
and
\begin{eqnarray}\label{eq:mean_rDt_NB1}
&&E_{\text{post}}(\log(f_{\text{NB1}}(N_{ti},\xi_{ti}\mid \wkt,\lambda_t)))\nonumber\\
&&=-2 \sum_{i=1}^{n_t} N_{ti} \left(\log(v_{ti})+\psi\left(\sum_{i=1}^{n_t} N_{t i}+\alpha\right)-\log\left(\ji{\sum_{i=1}^{n_t}} \xi_{ti}v_{t i}+\beta\right)\right)
 +2  \left(\frac{\sum_{i=1}^{n_t} N_{t i}+\alpha}{\sum_{i=1}^{n_t} \xi_{ti}v_{t i}+\beta}\right) \sum_{i=1}^{n_t} \xi_{t i}v_{ti}\nonumber\\
&&\qquad
+2 \sum_{i=1}^{n_t} \left(\log \left(N_{ti} !\right) 
-\left(\wkt+N_{ti}-1\right) \log \left(\xi_{ti}\right)+\xi_{ti} \wkt\right) +2(\log(\Gamma(\wkt))-\wkt \log (\wkt)).
\end{eqnarray}
Therefore, a direct calculation shows that the effective number of parameter\yz{s} for 
\ji{terminal} node $t$ is given by
$$
\begin{aligned}
r_{D{t}}=1+2 \left(\log \left(\sum_{i=1}^{n_t} N_{t i}+\alpha\right)-\psi\left(\sum_{i=1}^{n_t} N_{t i}+\alpha\right)\right) \ji{\sum_{i=1}^{n_t} N_{t i}},
\end{aligned}
$$
and thus
$$
\begin{aligned}
\text{DIC}_t &=D\left(\overline{\lambda_t}\right)+2 r_{Dt} \\
&=-2 \sum_{i=1}^{n_t} N_{ti} \left(\log(v_{ti})+\log\left(\sum_{i=1}^{n_t} N_{t i}+\alpha\right)-\log\left(\sum_{i=1}^{n_t} \xi_{ti}v_{t i}+\beta\right)\right)+2 \left(\frac{\sum_{i=1}^{n_t} N_{t i}+\alpha}{\sum_{i=1}^{n_t} \xi_{ti}v_{t i}+\beta}\right) \sum_{i=1}^{n_t} \xi_{t i}v_{ti} \nonumber
\\
& \ \ \ \ +2 \sum_{i=1}^{n_t} \left(\log \left(N_{ti} !\right) 
-\left(\wkt+N_{ti}-1\right) \log \left(\xi_{ti}\right)+\xi_{ti} \wkt\right)+2(\log(\Gamma(\wkt))-\wkt \log (\wkt)) \\
&\ \ \ \ +2+4 \left(\log \left(\sum_{i=1}^{n_t} N_{t i}+\alpha\right)-\psi\left(\sum_{i=1}^{n_t} N_{t i}+\alpha\right)\right) \ji{\sum_{i=1}^{n_t} N_{t i}}.
\end{aligned}
$$

\subsubsection{Negative binomial model 2 (NB2)}
We  now consider another  parameterization of the \ji{NB} distribution, see, e.g., \cite{lee2020delta, WuthrichMerz2022b}. Now, for terminal node $t$, 
\begin{eqnarray}\label{eq:NB2_dens}
    f_{\text{NB2}}(m\mid\kappa_t,\lambda_t) &=&P(N_{ti}=m\mid\kappa_t,\lambda_t)\nonumber\\
&=&\frac{\Gamma(m+\kappa_t v_{ti})}{\Gamma(\kappa_t v_{ti}) m!} \left(\frac{\kappa_t}{\kappa_t+\lambda_t}\right)^{\kappa_t v_{ti}} \left(\frac{\lambda_t}{\kappa_t+\lambda_t}\right)^{m},\quad \ji{m=0,1,\ldots.}
\end{eqnarray}
It is easy to show that the mean of $N_{ti}$ is the same as in \eqref{eq:N_mean}, but
 the variance becomes
\begin{eqnarray}\label{eq:N2_var}
\text{Var}(N_{ti}\mid \kappa_t,\lambda_t)= 
\lambda_t v_{ti}  \left(1+\frac{ \lambda_t}{\kappa_t}\right).
\end{eqnarray}
This formulation yields a fixed over-dispersion of size ${ \lambda_t}/{\kappa_t}$ which does not depend on the exposure $v_{ti}$, and thus it is sometimes preferred (see \cite{WuthrichMerz2022b}) and has been judged as more effective for real insurance data analysis (see \cite{lee2020delta}). 
 
\yz{We use the same way to deal with $\kappa_{t}$ and $\lambda_{t}$  as in the previous subsection.}
Using the same approach as Chapter 2 of \cite{Wuthrich2022}, we can estimate the parameter $\kappa_t$ as follows
\begin{equation}
\label{eq:NB2_kapp}
\widehat{\kappa}_t = \frac{\widehat{\lambda}_t^2 }{\widehat{V}_t^2 - \widehat{\lambda}_t},
\end{equation}
where \yz{$\widehat{V}_t^2$ and $\widehat{\lambda}_t$ are given in \eqref{eq:Vlambda}.}
Note that this parameterization offers a simpler estimation for $\wkt$, and that
$\widehat{\lambda}_t$ is a minimal variance  \yz{estimator}; see \cite{Wuthrich2022}.

Similarly as before, we can define a data augmented likelihood for the $i$-th data \ji{instance in terminal} node $t$ as
\begin{equation}
\label{nb2_augment}
f_{\text{NB2}}\left(N_{ti}, \xi_{ti} \mid \widehat{\kappa}_t,\lambda_{t}\right)=\left(\lambda_{t} v_{ti}\right)^{N_{ti}} e^{-\xi_{ti} \lambda_{t}v_{ti}}  \frac{( \wkt v_{ti})^{\wkt v_{ti}} \xi_{ti}^{\wkt v_{ti}+N_{ti}-1} e^{-\xi_{ti} \wkt v_{ti} }}{\Gamma(\wkt v_{ti}) N_{ti} !}.
\end{equation}
Further, we see that $\xi_{ti}$, given data $N_{ti}$ and parameters, has a gamma distribution, i.e.,
\begin{equation}\label{eq:xi_post2}
    \xi_{ti}\mid N_{ti}, \widehat{\kappa}_t,\lambda_{t}\  \sim \  \text{Gamma}\left(\wkt v_{ti}+N_{ti}, \wkt v_{ti}+\lambda_t v_{ti} \right).
\end{equation}

Given the data augmented likelihood in \eqref{nb2_augment}, the estimated parameter $\wkt$ \ji{using \eqref{eq:NB2_kapp}}, and a conjugate gamma prior for $\lambda_{t}$ 
with   hyper-parameters   $\alpha, \beta>0$,
we can derive the integrated augmented likelihood for  \yz{terminal} node $t$ as follows
\begin{equation}
\begin{aligned}
\label{nb2_integrated}
&p_{\text{NB2}}\left(\vk{N}_{t}, \vk{\xi}_t\mid \vk{X}_{t}, \vk{v}_t, \wkt\right) \\
&=\frac{ \beta^\alpha}{ \Gamma(\alpha)} \prod_{i=1}^{n_{t}}  \left[\frac{(\wkt v_{ti})^{\wkt v_{ti}} v_{ti}^{N_{ti}}}{\Gamma(\wkt v_{ti}) N_{ti} !} \xi_{ti}^{\wkt v_{ti}+N_{ti}-1} e^{-\xi_{ti} \wkt v_{ti}}\right]
\frac{\Gamma(\sum_{i=1}^{n_{t}} N_{ti}+\alpha)}{(\sum_{i=1}^{n_{t}}\xi_{ti}v_{ti}+\beta)^{\sum_{i=1}^{n_{t}} N_{ti}+\alpha}}.
\end{aligned}
\end{equation}
From the above we see that the posterior distribution of $\lambda_{t}$, given the augmented data  \yz{$(\vk{N}_t, \vk{\xi}_t)$}, is given by
\yz{$$
\lambda_{t} \mid \vk{N}_t, \vk{\xi}_t \ \sim \ \text{Gamma}\left(\sum_{i=1}^{n_{t}} N_{ti}+\alpha,\sum_{i=1}^{n_{t}}\xi_{ti}v_{ti}+\beta\right).
$$}
The integrated augmented likelihood for the tree $\mathcal{T}$ is thus given by
\begin{equation}\label{eq:NB2_whole}
    p_{\text{NB2}}\left(\vk{N}, \vk{\xi} \mid\vk{X},\vk{v},\widehat{\vk{\kappa}}, \mathcal{T} \right)=\prod_{t=1}^{b} p_{\text{NB2}}\left(\vk{N}_{t}, \vk{\xi}_t\mid \vk{X}_{t}, \vk{v}_t, \wkt\right).
\end{equation}
Now, we discuss the DIC$_t$ for terminal node $t$ of this tree. Similarly, as in the previous subsection, we can easily check that 
$$
\begin{aligned}
&\text{DIC}_t =D\left(\overline{\lambda_t}\right)+2 r_{Dt} \\
&=-2 \sum_{i=1}^{n_t} N_{ti} \left(\log(v_{ti})+\log\left(\sum_{i=1}^{n_t} N_{t i}+\alpha\right)-\log\left(\sum_{i=1}^{n_t} \xi_{ti}v_{t i}+\beta\right)\right)+2  \left(\frac{\sum_{i=1}^{n_t} N_{t i}+\alpha}{\sum_{i=1}^{n_t} \xi_{ti}v_{t i}+\beta}\right) \sum_{i=1}^{n_t} \xi_{t i}v_{ti}\nonumber
\\
&\ \ \ \ +2 \sum_{i=1}^{n_t} \left(\log(\Gamma(\wkt v_{ti}))+\log \left(N_{ti} !\right) 
-\wkt v_{ti} \log (\wkt v_{ti})-\left(\wkt v_{ti}+N_{ti}-1\right) \log \left(\xi_{ti}\right)+\xi_{ti} \wkt v_{ti}\right) \\
&\ \ \ \ +2+4 \left(\log \left(\sum_{i=1}^{n_t} N_{t i}+\alpha\right)-\psi\left(\sum_{i=1}^{n_t} N_{t i}+\alpha\right)\right) \sum_{i=1}^{n_t} N_{t i}.
\end{aligned}
$$
For the above two \ji{NB} models, 
the DIC of tree $\mathcal{T}$ is \yz{obtained by using \eqref{eq:DIC}}.


With the above formulas derived in the two subsections for \ji{NB} models, we can  use the three-step approach proposed in Section \ref{Sec:post_pre}, together with Algorithm \ref{Alg:NB}, to search for an optimal tree and then obtain predictions for new data.

\begin{algorithm}
	\caption{One step of the MCMC algorithm for \yz{the} \ji{NB} BCART parameterized by $(\vk{\kappa}, \vk{\lambda}, \mathcal{T})$ using data augmentation}
	\hspace*{0.02in} {\bf Input:}
	Data $(\vk{X},\vk{v}, \vk{N})$ and current values $\left(\widehat{\vk{\kappa}}^{(m)}, \vk{\lambda}^{(m)}, \mathcal{T}^{(m)},\vk{\xi}^{(m)}\right)$ \\
	\hspace*{0.3in} {\bf 1:}
	Generate a candidate value \(\mathcal{T}^{*}\) with probability distribution \(q\left(\mT^{(m)}, \mT^{*}\right)\)\\
        \hspace*{0.3in} {\bf 2:}
	Estimate $\widehat{\vk{\kappa}}^{(m)}$, using \eqref{eq:NB1_kapp} (or \eqref{eq:NB2_kapp})\\
        \hspace*{0.3in} {\bf 3:}
	Sample $\vk{\xi}^{(m+1)} \sim p(\vk{\xi}\mid \vk{X}, \vk{v},\vk{N}, \widehat{\vk{\kappa}}^{(m+1)}, \vk{\lambda}^{(m)}, \mathcal{T}^{(m)})$, using \eqref{eq:xi_post} (or \eqref{eq:xi_post2})\\
	\hspace*{0.3in} {\bf 4:}
	Set the acceptance ratio $\alpha\left(\mT^{(m)}, \mT^{*}\right)=\min \left\{\frac{q\left(\mT^{*}, \mT^{(m)}\right)}{q\left(\mT^{(m)}, \mT^{*}\right)} \frac{p_{\text{NB}}\left(\vk{N},\vk{\xi}^{(m+1)} \mid \vk{X}, \vk{v},\widehat{\vk{\kappa}}^{(m+1)},\mT^{*}\right)}{p_{\text{NB}}\left(\vk{N},\vk{\xi}^{(m)} \mid \vk{X}, \vk{v},\widehat{\vk{\kappa}}^{(m)}, \mT^{(m)}\right)}
\frac{p\left(\mT^{*}\right)}{p\left(\mT^{(m)}\right)}, 1\right\}$\\
	\hspace*{0.3in} {\bf 5:}
	Update \(\mT^{(m+1)}=\mT^{*}\) with probability  $\alpha\left(\mT^{(m)}, \mT^{*}\right)$, otherwise, set $\mT^{(m+1)}=\mT^{(m)}$\\
	\hspace*{0.3in} {\bf 6:}
	Sample $\vk\lambda^{(m+1)} \sim\text{Gamma}\left(\sum_{i=1}^{n_{t}} N_{ti}+\alpha,\sum_{i=1}^{n_{t}}\xi_{ti}^{(m+1)}v_{ti}+\beta\right)$\\
 \hspace*{0.02in} {\bf Output:} 
	New values $\left(\widehat{\vk{\kappa}}^{(m+1)}, \vk{\lambda}^{(m+1)}, \mathcal{T}^{(m+1)},\vk{\xi}^{(m+1)}\right)$
\label{Alg:NB}
\end{algorithm}

\begin{remark}
(a). In step 4 of Algorithm \ref{Alg:NB}, $p_{\text{NB}}$ should be understood as either $p_{\text{NB1}}$ or $p_{\text{NB2}}$. Similar to Algorithms \ref{Alg:1} and \ref{Alg:2}, the sampling steps in Algorithm \ref{Alg:NB} should be done when necessary.

(b).  It is worth noting that our way of dealing with the parameter $\kappa$ is different from that in \cite{murray2021log} where a single $\kappa$ is sampled from a distribution and used for all terminal nodes. It turns out that \ji{that} way of dealing with $\kappa$ cannot give us good estimates in our simulation examples, whereas our way of first estimating $\kappa$ \ji{using moment matching method} for each node can give good estimates.

(c). There are other ways to parameterize the \ji{NB} distribution, see, e.g., \cite{zhou2012lognormal}. However, it looks that  these ways are normally discussed when \yz{there is no exposure involved}, so we will not cover them here.
\end{remark}

\subsection{Zero-Inflated Poisson models}\label{ZIP}

Insurance claims data \ji{normally involves a large volume of zeros}. Many policy-holders incur no claims, which does not necessarily mean that they were involved in no accidents, but they are probably less risky. In this section, \ji{depending on how the exposure is embedded in the model we discuss two ZIP models to better reflect the excessive zeros}, see, e.g., \cite{lee2021addressing}.  


\subsubsection{Zero-Inflated Poisson model 1 (ZIP1)}
 For terminal node $t$, we use the following ZIP distribution by embedding the exposure into the Poisson part (see \cite{murray2021log}) 
\begin{eqnarray}\label{eq:f_ZIP1}
f_{\text{ZIP1}}\left(m\mid \mu_{t}, \lambda_{t}\right)&=& \begin{cases} \dfrac{1}{1+\mu_t}+ \dfrac{\mu_t}{1+\mu_t} f_P(0\mid \lambda_t) & m=0, \\[10pt] 
\dfrac{\mu_t}{1+\mu_t} f_P(m\mid \lambda_t)& m=1,2, \ldots, \end{cases}\nonumber\\
&=&  \dfrac{1}{1+\mu_t} I_{(m=0)}+ \frac{\mu_t}{1+\mu_t} f_P(m\mid \lambda_t),\ \ m=0,1,2, \ldots,
\end{eqnarray}
where $f_\text{P}(m\mid \lambda_t)$ is given as in \eqref{eq:f_P}, and $\frac{1}{1+\mu_t}\in(0,1)$ is the probability that a zero is due to the point mass component. Note that for computational simplicity we  \yz{consider} a model with two parameters rather than three as in \cite{murray2021log}. 

Similar to  the \ji{NB} model, a data augmentation scheme is needed for the ZIP model. To this end, we
introduce two latent variables $\vk{\phi}_t=(\phi_{t1},\phi_{t2},\ldots,\phi_{tn_t}) \in(0, \infty)^{n_t}$ and $\vk{\delta}_t=(\delta_{t1},\delta_{t2}, \ldots, \delta_{tn_t}) \in\{0,1\}^{n_t}$, and define the data augmented likelihood for the $i$-th data instance in terminal 
\yz{node $t$} by
\begin{equation}
\label{f_ZIP1_aug}
\begin{aligned}
f_{\text{ZIP1}}\left(N_{t i}, \delta_{t i}, \phi_{t i} \mid \mu_t,  \lambda_{t}\right) =
e^{-\phi_{ti}(1+\mu_{t})} \left( \frac{\mu_t\left(\lambda_{ t} v_{t i}\right)^{ N_{t i}} }{N_{t i} !} e^{-\lambda_{t} v_{t i}} \right)^{\delta_{t i}},  
\end{aligned}
\end{equation}
where the support of the function $f_{\text{ZIP1}}$ is $\left(\{0\}\times\{0,1\}\times(0,\infty)\right) \cup \left(\mathbb{N}\times\{1\}\times(0,\infty)\right)$. This means that we impose $\delta_{ti}=1$ when $N_{ti}\in \mathbb{N}$ (i.e., $N_{ti}\neq 0$). 
It can be shown that \eqref{eq:f_ZIP1} is the marginal distribution of the above augmented distribution; see \cite{murray2021log} for more details. By conditional arguments, we can also check that $\delta_{ti}$, given 
data $\phi_{ti}, N_{ti}=0$ and parameters, has a Bernoulli distribution, i.e.,
\begin{equation}\label{eq:delta_post}
    \delta_{ti} \mid\phi_{ti}, N_{ti}=0, \mu_t, \lambda_{t}\  \sim \  \text{Bern}\left(\frac{\mu_t e^{-\lambda_t v_{ti}}}{1+\mu_t e^{-\lambda_t v_{ti}}}\right),
\end{equation}
and $\delta_{ti}=1$, given $N_{ti}>0$.
Furthermore, $\phi_{ti}$, given data $\delta_{ti}, N_{ti}$ and parameters, has an exponential distribution, i.e.,
\begin{equation}\label{eq:phi_post}
    \phi_{ti} \mid\delta_{ti}, N_{ti}, \mu_t, \lambda_{t}\  \sim \  \text{Exp}\left(1+\mu_t\right).
\end{equation}
\lji{It is noted   that the augmented likelihood $f_{\text{ZIP1}}$ in \eqref{f_ZIP1_aug} can actually be factorized as two gamma-type functions parameterized by $\mu_t$ and $\lambda_t$ respectively. This observation motivates us to} assume independent conjugate gamma priors for $\mu_t$ and $\lambda_t$ \yz{with hyper-parameters   $\alpha_i, \beta_i>0$, $i=1,2$ (cf. \eqref{eq:p_lambda})}. With 
these gamma priors, we can derive the integrated augmented likelihood for terminal node $t$ as follows
\begin{equation}
\begin{aligned}
\label{eq:ZIP1_integrated}
p_{\text{ZIP1}}\left(\vk{N}_{t}, \vk{\delta}_t,\vk{\phi}_t\mid \vk{X}_{t}, \vk{v}_t \right) 
&=\int_0^\infty \int_0^\infty f_{\text{ZIP1}}\left(\vk{N}_t, \vk{\delta}_t, \vk{\phi}_t \mid \mu_t,\lambda_{t}  \right) p(\mu_t) p(\lambda_{t} )d\mu_t d \lambda_{t} \\
&= \frac{\beta_1^{\alpha_1}}{\Gamma\left(\alpha_1\right)} \frac{\beta_2^{\alpha_2}}{\Gamma\left(\alpha_2\right)} \prod_{i=1}^{n_t} \left(e^{-\phi_{t i}} v_{t i}^{\delta_{t i} N_{t i}}\left({N_{t i} !}\right)^{-\delta_{t i}} 
\right) \\
&\qquad \times \frac{\Gamma\left(\sum_{i=1}^{n_t} \delta_{t i}+\alpha_1\right)}{\left(\sum_{i=1}^{n_t} \phi_{t i}+\beta_1\right)^{\sum_{i=1}^{n_t} \delta_{t i}+\alpha_1}} \frac{\Gamma\left(\sum_{i=1}^{n_t} \delta_{t i} N_{t i}+\alpha_2\right)}{\left(\sum_{i=1}^{n_t} \delta_{t i} v_{t i}+\beta_2\right)^{\sum_{i=1}^{n_t} \delta_{t i} N_{t i}+\alpha_2}}.
\end{aligned}
\end{equation}
Moreover, from the above we see that the posterior distribution\yz{s} of $\mu_t, \lambda_{t}$ given the augmented data  \yz{$(\vk{N}_t, \vk{\delta}_t, \vk{\phi}_t)$}  \yz{are} given by
\begin{eqnarray*}
\mu_{t} \mid \vk{N}_t, \vk{\delta}_t, \vk{\phi}_t &\sim& \text{Gamma}\left(\sum_{i=1}^{n_{t}}\delta_{ti}+\alpha_{1},\sum_{i=1}^{n_{t}}\phi_{ti}+\beta_{1}\right),\\
\lambda_{t} \mid \vk{N}_t, \vk{\delta}_t, \vk{\phi}_t &\sim& \text{Gamma}\left(\sum_{i=1}^{n_{t}}\delta_{ti}N_{ti}+\alpha_{2},\sum_{i=1}^{n_{t}}\delta_{ti}v_{ti}+\beta_{2}\right).
\end{eqnarray*}
The integrated augmented likelihood for the tree $\mathcal{T}$ is thus given by


\begin{equation}\label{eq:ZIP1_whole}
   p_{\text{ZIP1}}\left(\vk{N}, \vk{\delta},\vk{\phi}\mid \vk{X}, \vk{v},  \mathcal{T} \right)=\prod_{t=1}^{b} p_{\text{ZIP1}}\left(\vk{N}_{t}, \vk{\delta}_t,\vk{\phi}_t\mid \vk{X}_{t}, \vk{v}_t \right).
\end{equation}
Now, we discuss the DIC for this tree which can be derived as a special case of \eqref{eq:DICt} with $\vk{\theta}_t=(\mu_t, \lambda_t)$. To this end, we first focus on the DIC$_t$ of terminal node $t$. It follows that
\begin{eqnarray}\label{eq:ZIP1_D_mu_lamb}
    D\left(\overline{\mu_{ t}}, \overline{\lambda_{t}}\right)&=&-2 \log f_{\text{ZIP1}}(\vk{N}_t\mid \overline{\mu_t}, \overline{\lambda_t})\nonumber\\
    &=&-2 \sum_{i=1}^{n_t} \log\left(\frac{1}{1+\overline{\mu_t}}I_{(N_{ti}=0)} +\frac{\overline{\mu_t}}{1+\overline{\mu_t}}\frac{(\overline{\lambda_t} v_{ti})^{N_{ti}}}{N_{ti}!} e^{-\overline{\lambda_t}v_{ti}} \right),
\end{eqnarray}
where
\begin{equation*}
    \overline{\mu_t}= \frac{\sum_{i=1}^{n_t} \delta_{t i}+\alpha_1}{\sum_{i=1}^{n_t} \phi_{t i}+\beta_1}, \qquad \overline{\lambda_t}= \frac{\sum_{i=1}^{n_t} \delta_{t i} N_{t i}+\alpha_2}{\sum_{i=1}^{n_t} \delta_{t i} v_{t i}+\beta_2}.
\end{equation*}

Next, since
\begin{eqnarray*}
 && 
 \log f_{\text{ZIP1}}\left(\vk{N}_{t}, \vk{\delta}_t,\vk{\phi}_t\mid \mu_t, \lambda_t \right)\\
 && 
 =\sum_{i=1}^{n_t} \left(-\phi_{t i}(1+\mu_{t})+\delta_{t i} \log \left(\mu_{ t}\right)+\delta_{t i} N_{t i} \log \left(\lambda_{ t} v_{t i}\right)-\delta_{t i} \lambda_{ t} v_{t i}-\delta_{t i} \log \left(N_{t i} !\right)\right),
\end{eqnarray*}
 we can derive that
\begin{eqnarray}\label{eq:ZIP1_pDt}
    q_{Dt} &=&- 2 E_{\text{post}}( \log f_{\text{ZIP1}}\left(\vk{N}_{t}, \vk{\delta}_t,\vk{\phi}_t\mid \mu_t, \lambda_t \right)) +2 \log  f_{\text{ZIP1}}\left(\vk{N}_{t}, \vk{\delta}_t,\vk{\phi}_t\mid \overline{\mu_t}, \overline{\lambda_t }\right)
    \nonumber\\ 
&=&2 \left(\log \left(\sum_{i=1}^{n_t} \delta_{t i}+\alpha_1\right)-\psi\left(\sum_{i=1}^{n_t} \delta_{t i}+\alpha_1\right)\right) \ji{\sum_{i=1}^{n_t} \delta_{t i}} \nonumber\\
&&\qquad + 2 \left(\log \left(\sum_{i=1}^{n_t} \delta_{t i} N_{t i}+\alpha_2\right)-\psi\left( \sum_{i=1}^{n_t} \delta_{t i} N_{t i}+\alpha_2\right)\right) \ji{\sum_{i=1}^{n_t} \delta_{t i} N_{t i}}.
\end{eqnarray}
\ji{Therefore, $\text{DIC}_t$ can be obtained from \eqref{eq:ZIP1_D_mu_lamb} and \eqref{eq:ZIP1_pDt} as}
\begin{eqnarray*}
\text{DIC}_t =D\left(\overline{\mu_{ t}}, \overline{\lambda_{ t}}\right)+2 q_{Dt}. 
\end{eqnarray*}

\subsubsection{Zero-Inflated Poisson model 2 (ZIP2)}
For terminal node $t$, we use the following ZIP distribution by embedding the exposure into the zero mass part (see \cite{lee2021addressing})
\begin{equation}\label{eq:f_ZIP2}
f_{\text{ZIP2}}\left(m\mid \mu_{t}, \lambda_{t}\right)=
\begin{cases} \dfrac{1}{1+\mu_t v_{ti}}+ \dfrac{\mu_t v_{ti}}{1+\mu_t v_{ti}} e^{- \lambda_t} & m=0, \\[10pt] 
\dfrac{\mu_t v_{ti}}{1+\mu_t v_{ti}}\dfrac{\lambda_t^{N_{ti}}}{N_{ti}!} e^{-\lambda_t}& m=1,2, \ldots,\end{cases} 
\end{equation}
where 
$\frac{1}{1+\mu_t v_{ti}}\in(0,1)$ is the probability that a zero is due to the point mass component. This formulation stems from an  intuitive inverse relationship between the exposure and the probability of zero mass.  This way of embedding exposure has been justified to be more effective in \cite{lee2021addressing}. 

Similar to before, we
introduce two latent variables $\vk{\phi}_t=(\phi_{t1},\phi_{t2},\ldots,\phi_{tn_t}) \in(0, \infty)^{n_t}$ and $\vk{\delta}_t=(\delta_{t1},\delta_{t2}, \ldots, \delta_{tn_t}) \in\{0,1\}^{n_t}$, and define the data augmented likelihood for the $i$-th data instance 
\yz{in terminal node $t$} as
\begin{equation}
\label{f_ZIP2_aug}
\begin{aligned}
f_{\text{ZIP2}}\left(N_{t i}, \delta_{t i}, \phi_{t i} \mid \mu_t,  \lambda_{t}\right) =
  e^{-\phi_{t i}\left(1+ \mu_t v_{ti}\right)}  \left( \frac{\mu_t v_{ti} \lambda_{ t} ^{ N_{t i}} }{N_{t i} !} e^{-\lambda_{t} } \right)^{\delta_{t i}},  
\end{aligned}
\end{equation}
where the support of the function $f_{\text{ZIP2}}$ is $\left(\{0\}\times\{0,1\}\times(0,\infty)\right) \cup \left(\mathbb{N}\times\{1\}\times(0,\infty)\right)$. 
By conditional arguments, we can also check that $\delta_{ti}$, given 
data $\phi_{ti}, N_{ti}=0$ and parameters, has a Bernoulli distribution, i.e.,
\begin{equation}\label{eq:delta_post_2}
    \delta_{ti} \mid\phi_{ti}, N_{ti}=0, \mu_t, \lambda_{t}\  \sim \  \text{Bern}\left(\frac{\mu_t v_{ti} e^{-\lambda_t }}{1+ \mu_t v_{ti} e^{-\lambda_t }}\right),
\end{equation}
and $\delta_{ti}=1$, given $N_{ti}>0$. 
Furthermore, $\phi_{ti}$, given data $\delta_{ti}, N_{ti}$ and parameters, has an exponential distribution, i.e.,
\begin{equation}\label{eq:phi_post_2}
    \phi_{ti} \mid\delta_{ti}, N_{ti}, \mu_t, \lambda_{t}\  \sim \  \text{Exp}\left(1+\mu_tv_{ti}\right).
\end{equation}
 As previously, we assume  independent conjugate gamma priors \yz{for $\mu_t$ and $\lambda_t$ with hyper-parameters   $\alpha_i, \beta_i>0$, $i=1,2$. 
 } Given the data augmented likelihood in \eqref{f_ZIP2_aug} and the above gamma priors, we can obtain the integrated augmented likelihood for terminal node $t$ as follows
\begin{equation}
\begin{aligned}
\label{eq:ZIP2_integrated}
p_{\text{ZIP2}}\left(\vk{N}_{t}, \vk{\delta}_t,\vk{\phi}_t\mid \vk{X}_{t}, \vk{v}_t \right) 
&= \frac{\beta_1^{\alpha_1}}{\Gamma\left(\alpha_1\right)} \frac{\beta_2^{\alpha_2}}{\Gamma\left(\alpha_2\right)} \prod_{i=1}^{n_t} \left(e^{-\phi_{t i}} \left(\frac{v_{ti}}{{N_{t i} !} }\right)^{\delta_{t i}} 
\right) \\
&\qquad \times \frac{\Gamma\left(\sum_{i=1}^{n_t} \delta_{t i}+\alpha_1\right)}{\left(\sum_{i=1}^{n_t} \phi_{t i} v_{ti}+\beta_1\right)^{\sum_{i=1}^{n_t} \delta_{t i}+\alpha_1}} \frac{\Gamma\left(\sum_{i=1}^{n_t} \delta_{t i} N_{t i}+\alpha_2\right)}{\left(\sum_{i=1}^{n_t} \delta_{t i} +\beta_2\right)^{\sum_{i=1}^{n_t} \delta_{t i} N_{t i}+\alpha_2}}.
\end{aligned}
\end{equation}
Moreover, from the above we see that the posterior distribution\yz{s} of $\mu_t, \lambda_{t}$ given the augmented data  \yz{$(\vk{N}_t, \vk{\delta}_t, \vk{\phi}_t)$} \yz{are} given by
\begin{eqnarray*}
\mu_{t} \mid \vk{N}_t, \vk{\delta}_t, \vk{\phi}_t &\sim& \text{Gamma}\left(\sum_{i=1}^{n_{t}}\delta_{ti}+\alpha_{1},\sum_{i=1}^{n_{t}}\phi_{ti}v_{ti}+\beta_{1}\right),\\
\lambda_{t} \mid \vk{N}_t, \vk{\delta}_t, \vk{\phi}_t &\sim& \text{Gamma}\left(\sum_{i=1}^{n_{t}}\delta_{ti}N_{ti}+\alpha_{2},\sum_{i=1}^{n_{t}}\delta_{ti}+\beta_{2}\right).
\end{eqnarray*}
The integrated augmented likelihood for the tree $\mathcal{T}$ is thus given by

\begin{equation}\label{eq:ZIP2_whole}
   p_{\text{ZIP2}}\left(\vk{N}, \vk{\delta},\vk{\phi}\mid \vk{X}, \vk{v},  \mathcal{T} \right)=\prod_{t=1}^{b} p_{\text{ZIP2}}\left(\vk{N}_{t}, \vk{\delta}_t,\vk{\phi}_t\mid \vk{X}_{t}, \vk{v}_t,  \mathcal{T} \right).
\end{equation}
Now, we discuss 
the DIC$_t$ of terminal node $t$. It follows that
\begin{align}\label{eq:ZIP2_D_mu_lamb}
    D\left(\overline{\mu_{ t}}, \overline{\lambda_{t}}\right)&=-2 \log f_{\text{ZIP2}}(\vk{N}_t\mid \overline{\mu_t}, \overline{\lambda_t})\nonumber\\
    &=-2 \sum_{i=1}^{n_t} \log\left(\frac{1}{1+\overline{\mu_t} v_{ti}}I_{(N_{ti}=0)} +\frac{\overline{\mu_t} v_{ti}}{1+\overline{\mu_t} v_{ti}}\frac{\overline{\lambda_t} ^{N_{ti}}}{N_{ti}!} e^{-\overline{\lambda_t}} \right),
\end{align}
where
\begin{equation*}
    \overline{\mu_t}= \frac{\sum_{i=1}^{n_t} \delta_{t i}+\alpha_1}{\sum_{i=1}^{n_t} \phi_{t i} v_{ti}+\beta_1}, \qquad \overline{\lambda_t}= \frac{\sum_{i=1}^{n_t} \delta_{t i} N_{t i}+\alpha_2}{\sum_{i=1}^{n_t} \delta_{t i}+\beta_2}.
\end{equation*}
Next, since
\begin{eqnarray*}
 && \log f_{\text{ZIP2}}\left(\vk{N}_{t}, \vk{\delta}_t,\vk{\phi}_t\mid \mu_t, \lambda_t \right)\\
 && =\sum_{i=1}^{n_t} \left(-\phi_{t i}(1+\mu_tv_{ti})+\delta_{t i} \log \left(\mu_{ t}v_{ti}\right)+\delta_{t i} N_{t i} \log \left(\lambda_{ t} \right)-\delta_{t i} \lambda_{ t} -\delta_{t i} \log \left(N_{t i} !\right)\right),
\end{eqnarray*}
 we can derive the \lj{same expression for $q_{Dt}$ as in \eqref{eq:ZIP1_pDt}. 
 } Therefore, we obtain from \eqref{eq:ZIP2_D_mu_lamb} and \eqref{eq:ZIP1_pDt} that
 \COM{ 
\begin{eqnarray}\label{eq:ZIP2_pDt}
    p_{Dt}
&=&2 \sum_{i=1}^{n_t} \delta_{t i}\left(\log \left(\sum_{i=1}^{n_t} \delta_{t i}+\alpha_1\right)-\psi\left(\sum_{i=1}^{n_t} \delta_{t i}+\alpha_1\right)\right) \nonumber\\
&&+2 \sum_{i=1}^{n_t} \delta_{t i} N_{t i}\left(\log \left(\sum_{i=1}^{n_t} \delta_{t i} N_{t i}+\alpha_2\right)-\psi\left( \sum_{i=1}^{n_t} \delta_{t i} N_{t i}+\alpha_2\right)\right).
\end{eqnarray}
\yz{It should be called as $q_{Dt}$? Is it necessary to mention although different parameterization ways, effective number of parameters is the same..}
} 
\begin{align*}
\text{DIC}_t 
&=-2 \sum_{i=1}^{n_t} \log\left(\frac{1}{1+\overline{\mu_t} v_{ti}}I_{(N_{ti}=0)} +\frac{\overline{\mu_t} v_{ti}}{1+\overline{\mu_t} v_{ti}}\frac{\overline{\lambda_t} ^{N_{ti}}}{N_{ti}!} e^{-\overline{\lambda_t}} \right) \\
&\qquad +4 \left(\log \left(\sum_{i=1}^{n_t} \delta_{t i}+\alpha_1\right)-\psi\left(\sum_{i=1}^{n_t} \delta_{t i}+\alpha_1\right)\right) \sum_{i=1}^{n_t} \delta_{t i} \nonumber\\
&\qquad +4 \left(\log \left(\sum_{i=1}^{n_t} \delta_{t i} N_{t i}+\alpha_2\right)-\psi\left( \sum_{i=1}^{n_t} \delta_{t i} N_{t i}+\alpha_2\right)\right) \sum_{i=1}^{n_t} \delta_{t i} N_{t i}.
\end{align*}
For the above two ZIP models, 
the DIC of tree $\mathcal{T}$ is \yz{obtained using \eqref{eq:DIC}}.

With the formulas derived in the above two subsections for ZIP models, we can  use the three-step approach proposed in Section \ref{Sec:post_pre}, together with Algorithm \ref{Alg:2}, to search for an optimal tree and then obtain predictions for new data.

\begin{remark}
    There are other ways to deal with the data augmentation for ZIP models; see, e.g., \cite{tanner1987calculation, rodrigues2003bayesian, diebolt1994estimation} where only one latent variable is introduced. The models discussed therein with one latent variable should work more efficiently, \ji{but in their constructions no exposure is considered. Since involvement of exposure is one of the key features of insurance claims frequency analysis, 
    we had to introduce two latent variables for data augmentation to facilitate calculations. }
    
    \COM{The advantage is this method only introduces one latent variable when the exposure is not considered, which decreases the randomness a lot. However, even if the latent variable can be integrated out, it does not obtain an equation that equals to the data likelihood. Luckily, Diebolt \cite{diebolt1994estimation} showed that the two likelihoods (data likelihood and data augmented likelihood) can achieve convergence even if they are not equal. Recently, Murray \cite{murray2021log} proposed a new data augmentation way, and the data augmented likelihood in this method is equal to the data likelihood if the latent variables are integrated out. We use both of the two ways to check the performance and the latter one is better, therefore, the paper only gives the detailed calculation procedure for the second method below. ?? needs some changes...}
\end{remark}

\section{Simulation and real data analysis}\label{sec:sim}
In this section, we illustrate the efficiency of the  BCART models \ga{introduced}  in Section \ref{Sec:CF} by using  simulated data and a real insurance claims dataset. In the sequel, we use the abbreviation P-CART to denote CART for the Poisson model, 
and other abbreviations can be similarly understood (e.g., NB1-BCART denotes the BCART for \ji{\ji{NB}} model 1).

\subsection{Performance measures}
We first introduce some performance measures that will be used for prediction comparisons.
Suppose we \ji{have} obtained a tree with $b$ terminal nodes and \ji{the corresponding parameter estimates for $\vk{\theta}$} which we will use to obtain the prediction  $\widehat{N}_{i}$ given $\vk{x}_i, v_i$  
for a test data \ji{set} $(\vk{X}, \vk{v}, \vk{N})$ with $m$ observations. The number of test data in the terminal node $t$ is denoted by $m_t$, $t=1,\ldots,b.$
The performance measures used are as follows:

\begin{itemize}
    \item[{\bf M1:}] The residual sum of squares (RSS) is given by 
    $$ \text{RSS}(\vk{N})=\sum_{i=1}^{m} 
    (N_{i}-\widehat{N}_i)^2.$$ 


This measure is  commonly used for Gaussian-distributed data, but here we also use it for non-Gaussian data for comparison.
    \item[{\bf M2:}] 
    RSS based on a sub-portfolio  (i.e., \lj{those} instances in the same terminal node) level is given by 
    $$ \text{RSS}(\vk{N/v})=\sum_{t=1}^{b} \left(\frac{\sum_{i=1}^{m_{t}} N_{ti}}{\sum_{i=1}^{m_{t}} v_{ti}}-\widehat{y}_t\right)^2,$$
    where $\widehat{y}_t$ is the estimated frequency for the terminal node $t$, which is estimated by \eqref{eq:yhat}  \ji{assuming unit exposure}. More specifically, $\widehat{y}_t =\overline\lambda_t$ for Poisson and \ji{NB} models, and $\widehat{y}_t =\overline{\mu}_t\overline\lambda_t(1+\overline{\mu}_t)^{-1} $ for \yz{ZIP models}.
    This measure is preferred here as it takes account of  accuracy on a (sub-)portfolio level (i.e., balance property) other than an individual level. We refer to \cite{denuit2021autocalibration, wuthrich2020bias, wuthrich2022balance} for more details and discussions of the balance property that is required for insurance pricing.
    \item[{\bf M3:}] Negative log-likelihood (\ji{NLL}): This is calculated by \yz{using the assumed response distribution in the terminal node} with the estimated parameters. 
    It  represents the ex-ante belief of the underlying distribution of the data, and is thus a good measure for model comparison, see, e.g., \cite{lee2021addressing}.
    \item[{\bf M4:}] Discrepancy statistic (DS) \yz{(cf. \cite{naya2008comparison})}, is defined as a weighted version of RSS($\vk{N/v}$), given by
     $$ \text{DS}(\vk{N/v})=\sum_{t=1}^{b} \frac{1}{\widehat{\sigma}^2_t}\left(\frac{\sum_{i=1}^{m_{t}} N_{ti}}{\sum_{i=1}^{m_{t}} v_{ti}}-\widehat{y}_t\right)^2,$$
     where $\widehat{y}_t$ is the same as in M2, and $\widehat{\sigma}^2_t$ is the estimated variance 
     of frequency for terminal node $t$. More specifically, $\widehat{\sigma}^2_t =\overline\lambda_t$ for the Poisson model, $\widehat{\sigma}^2_t =\overline\lambda_t (1+\overline\lambda_t/\widehat{\kappa}_t)$ for  \ji{NB} models, \yz{and} $\widehat{\sigma}^2_t =\overline{\mu}_t\overline\lambda_t (1+\overline{\mu}_t+\overline\lambda_t)\left(1+\overline{\mu}_t\right)^{-2} $ for \yz{ZIP models}.
    \item[{\bf M5:}] Lift: Model lift indicates the ability to differentiate between low and high \ji{claims} frequency policy-holders. Sometimes it is called the “economic value” of the model.  A higher lift illustrates that the model is more capable of separating the extreme values from the average. We refer to \cite{henckaerts2021boosting, lee2020delta, lee2021addressing} \ji{and references therein} for further discussion on lift. We propose a way to calculate  lift for the tree model in the \yz{following} steps.
\begin{itemize}
    \item[Step 1:] Retrieve the predicted frequencies for terminal nodes, $\widehat{y}_t, t=1,\ldots,b,$ for the optimal tree obtained from the training procedure.
 \item[Step 2:] Set $\widehat{y}_{\min}=\min_{t=1}^b \widehat{y}_t$ and $\widehat{y}_{\max}=\max_{t=1}^b \widehat{y}_t$, which identify the least and most risky groups of policy-holders, respectively.
    \item[Step 3:] Use test data in the least and most risky groups/nodes to obtain their \ji{total sum of} exposures, say $v_{\min}$ and $v_{\max}$. 
    \item[Step 4:] If $v_{\min} \le v_{\max}$, then sort the data  using exposures in  descending order in the most risky group. Calculate the cumulative sums of the sorted exposures until \ji{the one equal or greater 
    than $v_{\min}$ is achieved} and then calculate the corresponding empirical frequency (i.e., ratio of sum of \yz{claim} numbers and sum of exposures) of these first data involved, say $\lambda_{\max |}^{(e)}$. The lift is defined as $L=\lambda_{\max |}^{(e)}/\lambda_{\min }^{(e)}$, where $\lambda_{\min }^{(e)}$ is the empirical frequency of the least risky group.

    [Similarly, If $v_{\min}>v_{\max}$, then sort the data  using exposures in  ascending order in the least risky group. Calculate the cumulative sums of the sorted exposures until the one equal or greater 
    than $v_{\max}$ is achieved and then calculate the corresponding empirical frequency of these first data involved, say $\lambda_{\min|}^{(e)}$. The lift is defined as $L=\lambda_{\max }^{(e)}/\lambda_{\min| }^{(e)}$, where $\lambda_{\max }^{(e)}$ is the empirical frequency of the most risky group.]
 
\end{itemize}

\end{itemize}

We remark that more performance measures and diagnostic approaches can be introduced following ideas in, e.g.,  \cite{lee2021addressing, lee2020delta, henckaerts2021boosting}. However, this is not the main focus of the present paper, so these 
are explored elsewhere.

\subsection{Simulation examples}

We will discuss three simulation examples, namely, Scenarios 1--3 below. 
Scenario 1 aims to illustrate that BCART can do really well for the \ji{chessboard data similar to}  Figure \ref{Fig:S1_data} for which CART cannot reasonably do anything. In addition, from this simulation study we \ji{also} see that BCART can do well with variable selection. In Scenario 2, we shall examine how different BCART models can capture the \ji{data over-dispersion.} In Scenario 3, we illustrate the effectiveness of ZIP-BCART models for data with exposures.

\subsubsection{Scenario 1: Poisson data with noise variables} \label{subsec:S1}

We simulate a data set $\{(\vk{x}_i,v_i,N_i)\}_{i=1}^n$ with $n=5,000$ independent observations. Here $v_{i} \sim \text{U}(0,1)$, $\vk{x}_i=(x_{i1}, 
\ldots, x_{i8})$, \ji{with \yao{independent components} $x_{i1}\sim \text{U}\{-3,-2,-1,1,2,3\}$, $x_{i2}\sim \text{N}(0,1)$, $x_{ik}\sim \text{U}(-1,1)$ for $k=3,4$, $x_{ik}\sim \text{N}(0,1)$ for $k=5,6$, and $x_{ik}\sim \text{U}\{-3,-2,-1,1,2,3\}$ for $k=7,8$.} Moreover, $N_i\sim \text{Poi}(\lambda \left(x_{i1},x_{i2}\right) v_i)$, where
 $$
	\lambda \left(x_{1},x_{2}\right)=\left\{\begin{array}{ll}
	1 & \text { if } x_{1}x_{2} \leq 0, \\
	7 & \text { if } x_{1}x_{2} > 0.
	\end{array}\right.
	$$
\yao{Obviously, the designed noise variables $x_{ik}, k=3,\ldots, 8$ are \yao{all} independent of the response $\vk{N}$.} 
We use P-BCART and P-CART for the above simulated data, where $x_{ik}, k=1,7,8$ are   treated as categorical. \yz{We have included both categorical and continuous variables as noise variables and as significant variables, which is a bit more general than the data shown in Figure \ref{Fig:S1_data}.} Note that the same conclusion can be drawn for numeric $x_{ik}, k=1,7,8$, but to better illustrate the effectiveness of the P-BCART we  choose to make them as characters (to increase the splitting possibilities of these variables). 

We first apply P-CART as implemented in the R package \texttt{rpart} \cite{R:rpart}. It is not surprising 
that P-CART is not able to give us any reasonable tree that can characterize the data, due to its greedy search nature.
The smallest tree (except the one with only a root node) that P-CART \lj{generated} has 25 terminal nodes and the tree found by using cross-validation has 31 terminal nodes. Obviously, both of them are much more complicated than the real model. Furthermore, in these two trees all the noise variables are used, which indicates that P-CART is sensitive to noise. 

\llj{Now we discuss the P-BCART applied to the data focusing preliminary on the effect of noise variables to the model. We simply set equal probabilities, i.e.,
P(Grow)$=0.2$, P(Prune)$=0.2$, P(Change1)$=0.2$, P(Change2)$=0.2$ and P(Swap)$=0.2$, for the tree proposals.} 
For the gamma prior of the Poisson intensities $\lambda_t$ we use $\alpha=3.2096$ and $\beta=0.8$ which are selected by keeping the relationship \ji{${\alpha}/{\beta}={\sum_{i=1}^n{N_{i}}}/{\sum_{i=1}^n{v_{i}}}$}. It is worth mentioning that the performance of the algorithm does not change much when choosing different pairs of $(\alpha, \beta)$ while keeping their ratio. We also observe the same in other simulation examples, so in the following we will not dwell on their selection.

In Table \ref{table_S1-1} we list the tuned hyper-parameters $\gamma, \rho$ in the first two columns for which the MCMC algorithms will converge to a region of trees with a certain number of terminal nodes listed (see Step 1 of Table \ref{table_SS}). For each fixed hyper-parameter $\gamma$ and $\rho$, we run 10000 iterations in the MCMC algorithm and take results after an initial burn-in period of 2000 iterations, after which the posterior probabilities of the tree structures have been settled for some time. This procedure is done with 3 restarts. The fourth column gives the total number of accepted trees after the burn-in period in the MCMC algorithms.
The last columns of Table \ref{table_S1-1} include the total number of times each variable is \yz{used} in the accepted trees. 
\llj{ 
We see from these columns that all noise variables have a very low selection rate, and 
as expected, the significant variables $x_1,x_2$ are dominating. 
Besides, at a first glance, it is infered 
that the noise variables $x_{3}$ and $x_{4}$ have a much lower selection rate than the other noise variables which is just because $x_{3}$ and $x_{4}$ are simulated using a distribution 
completely different from those of the significant variables.
However, when the experiment is run 10 times, we find that 
the average selection rates of all noise variables are almost the same independent of their distributions (see Table \ref{table_S1-1_10T}), which is consistent with the expectation. 
}

\begin{table}[!t] 
 \centering
 \caption{Total count each variable used amongst all accepted trees from the P-BCART MCMC algorithms (after burn-in period; \yao{equal probabilities for tree moves; one run \llj{with 3 restarts})}}
 \begin{tabular}{c|c|c|c|cccccccc}  

\toprule   

   $\gamma$ & $\rho$ & \ji{\# terminal nodes} & \ji{\#
   accepted trees} & $x_{1}$ & $x_{2}$ & $x_{3}$ & $x_{4}$ & $x_{5}$ & $x_{6}$ & $x_{7}$ & $x_{8}$\\  

\hline   

0.50 & 20 &2  & 342 & 197 & 156 & 1 & 0 & 2 & 5 & 4 & 4 \\

0.95 & 17 & 3 & 460  & 408 & 381 & 1 & 4 & 8 & 6 & 8 & 5
 \\

0.99 & 15 & 4 & 800 & 1261 & 1239 & 12 & 17 & 30 & 25 & 31 & 20 \\

0.99 & 12 & 5  & 652 & 1157 & 1126 & 9 & 7 & 18 & 15 & 16 & 20  \\

0.99 & 10 & 6  & 305 & 710 & 680 & 13 & 4 & 18 & 25 & 30 & 12 \\

0.99 & 6 & 7  & 318 &  825 & 809 & 3 & 8 & 15 & 23 & 14 & 9
 \\

0.99 & 5 &  8  & 210 & 681 & 647 & 2 & 10 & 7 & 13 & 18 & 5 \\

  \bottomrule  

\end{tabular}

 \label{table_S1-1}
\end{table}

\begin{table}[!t] 
 \centering
 \caption{\yao{Average frequency each variable used in all accepted trees from the P-BCART MCMC algorithms (after burn-in period; equal probabilities for tree moves; ten runs \llj{with 3 restarts})}}
 \begin{tabular}{c|ccccccccccc}  

\toprule   

\ji{\# terminal nodes} & $x_{1}$ & $x_{2}$ & $x_{3}$ & $x_{4}$ & $x_{5}$ & $x_{6}$ & $x_{7}$ & $x_{8}$\\  

\hline   

2  & 183 & 140 & 1 & 1 & 1 & 3 & 2 & 1 \\

 3 &  422 & 405 & 2 & 3 & 4 & 3 & 3 & 2
 \\

4 &  1242 & 1201 & 11 & 13 & 15 & 13 & 14 & 12 \\

 5 & 1207 & 1162 & 12 & 14 & 12 & 13 & 11 & 14  \\

 6 & 821 & 828 & 9 & 8 & 10 & 12 & 11 & 8 \\

 7  &  998 & 976 & 8 & 9 & 10 & 12 & 10 & 8
 \\

  8   & 847 & 795 & 7 & 9 & 9 & 11 & 10 & 8 \\

  \bottomrule  

\end{tabular}

 \label{table_S1-1_10T}
\end{table}

\begin{figure}[htbp]
	\centering
	\includegraphics[width=13cm]{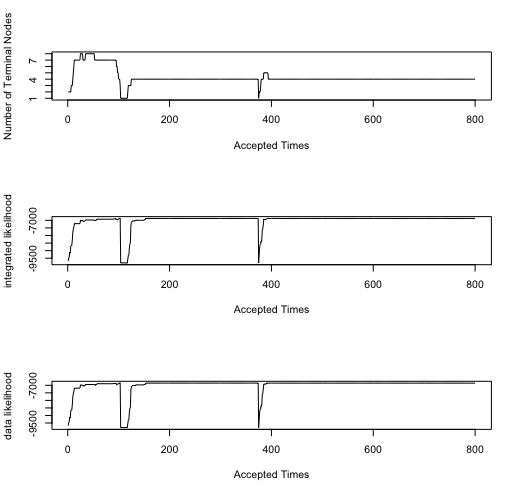}
	\caption{Trace plots \ji{from MCMC with 3 restarts }($\gamma=0.99,\rho=15$).}
    \label{Fig:S1_trace plots} 
\end{figure}

In Figure \ref{Fig:S1_trace plots}, we illustrate this procedure for $j=4$ \yz{(the same as that summarized in the third row of Table \ref{table_S1-1}),} with plots of the number of terminal nodes, the integrated likelihood $p_\text{P}(\vk{N}|\vk{X}, \vk{v},\vk{\mathcal{T}})$ and the \ji{data} likelihood $p_\text{P}(\vk{N}|\vk{X}, \vk{v}, \overline{\vk{\lambda}}, \vk{\mathcal{T}})$ of \ji{the accepted} trees. The observations are in line with those in \cite{chipman1998bayesian}; we see from the  likelihood plots  that the convergence of MCMC can be obtained relatively quickly. Interestingly, 
the optimal tree is not found in the first round of MCMC which got stuck in a local mode, but the restarts helped where in the second and the third round\yz{s} optimal trees can be found. Moreover, we see that there is no big difference shown in the plots of the integrated likelihood and the data likelihood.

\begin{figure}[htbp]
\centering
\includegraphics[width=10cm]{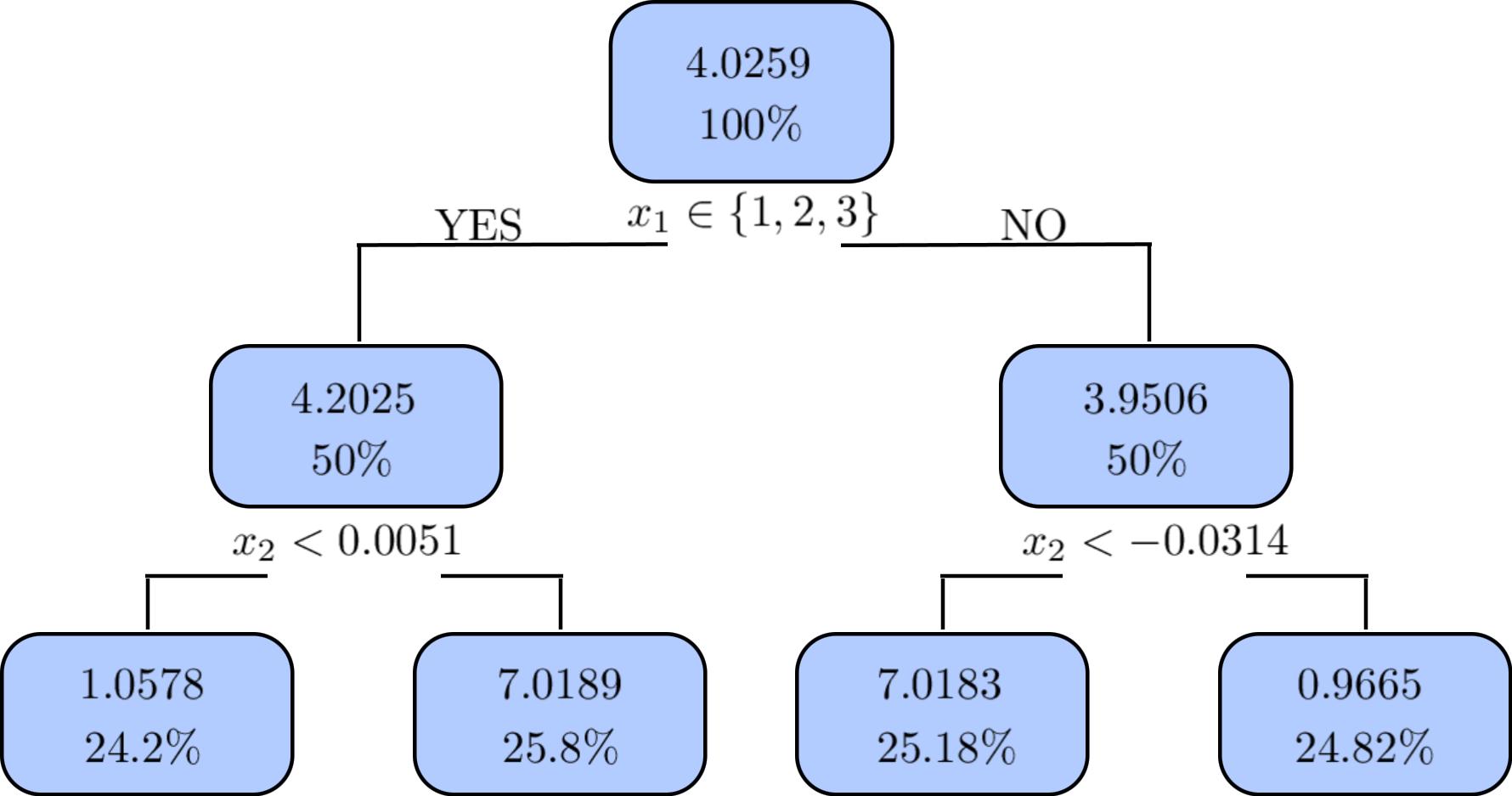}
	\caption{Optimal  P-BCART. Numbers at each node give the estimated value for the frequency parameter $\lambda_{t}$ and the percentage of observations.}
    \label{Fig:S1_optimal tree}
    
\end{figure}

Following Step 2 of Table \ref{table_SS}, for each $j=2,\ldots,8$, we select the optimal tree with maximum  \yz{data} likelihood $p_\text{P}(\vk{N}|\vk{X}, \vk{v}, \overline{\vk{\lambda}}, \vk{\mathcal{T}})$ from the convergence region. The variables used in these \yz{optimal} trees are listed in Table \ref{table_S1-2}, where we can see that none of these trees involves any of the noise variables. The values for the effective number of parameters $p_D$ reflect the number of parameters in the tree if a flat prior for $\lambda_t$ is used. Furthermore,  we list the DIC for these trees \ji{in} the last column of Table  \ref{table_S1-2}. \ji{Following Step 3 of Table \ref{table_SS}} we conclude that the selected optimal tree is the one with 4 terminal nodes which is illustrated in  Figure \ref{Fig:S1_optimal tree}. We see that this tree is \yz{close to} \yz{a true} optimal one with the almost correct topology and accurate parameter estimates. 

\begin{table}[!t] 

 \centering
  \caption{Number of times each variable used in each chosen optimal tree and the corresponding \yz{$p_D$ and} DIC (after burn-in period; \yao{equal probabilities for tree moves; one run} \llj{with 3 restarts}). Bold font indicates DIC selected model.}
 \begin{tabular}{c|cccccccc|c|c}  

\toprule   

   \ji{\# terminal nodes} & $x_{1}$ & $x_{2}$ & $x_{3}$ & $x_{4}$ & $x_{5}$ & $x_{6}$ & $x_{7}$ & $x_{8}$ & $p_D$ & DIC\\  

\hline 

2 & 1 & 0 & 0 & 0 & 0 & 0 & 0 & 0 & 2.00 & 
14221 \\

3 & 1 & 1 & 0 & 0 & 0 & 0 & 0 & 0 & 2.95 & 
14076 \\

4 & 1 & 2 & 0 & 0 & 0 & 0 & 0 & 0 & 3.97  & {\bf 
13526}\\

5 & 2 & 2 & 0 & 0 & 0 & 0 & 0 & 0 & 4.97 & 
13570\\

6 & 3 & 2 & 0 & 0 & 0 & 0 & 0 & 0 & 5.93 & 
13629 \\

7 & 3 & 3 & 0 & 0 & 0 & 0 & 0 & 0 & 6.91 & 
13678 \\
8 & 4 & 3 & 0 & 0 & 0 & 0 & 0 & 0 & 7.95 & 
13683 \\

  \bottomrule  

\end{tabular}

 \label{table_S1-2}
\end{table}

\llj{Using equal probabilities for the proposed tree moves, the above example provides detailed information about how to implement the three-step tree selection procedure in practice and illustrates the effectiveness of the method.  Next, we investigate 
which type of step (particularly, the Change and Swap moves) contributes more to the computational efficiency. To this end, we shall vary the probabilities of the Change and Swap moves, keeping the same probabilities for Grow and Prune moves at 0.2. Different experiments can be designed as in Table \ref{table_S1-1_VaryTreeMoves}.}

\begin{table}[!t] 
 \centering
 \caption{\yao{Four different experiments (E1--E4) for given probabilities of tree moves. In each case probability of Grow and Prune is fixed at $0.2$.}}
 \begin{tabular}{c|cccc}  

\toprule   

     & Change1 & Change2 & Swap \\  

\hline

E1 & 0 & 0.6 & 0 \\
E2 & 0 & 0.3 & 0.3\\
E3 & 0.3 & 0 & 0.3 \\
E4& 0.2 & 0.2 & 0.2 \\

  \bottomrule  

\end{tabular}\bigskip

 \label{table_S1-1_VaryTreeMoves}
 \centering
 \caption{\yao{Average iteration times \llj{to obtain an ``optimal'' tree (4 terminal nodes)} and accepted move rates  from the P-BCART MCMC algorithms (after burn-in period; ten runs \llj{with 3 restarts}). Experiments E1--E4 are described in Table \ref{table_S1-1_VaryTreeMoves}.}}
\begin{tabular}{c|ccccc}  
\toprule   
     & E1 & E2 & E3 & E4 \\  
\hline   
\yz{Average iteration times (s.d.)} & 3388 (168) & 2710 (187) & 2984 (177) & 2018 (161) \\
Acceptance rate of all moves & 3.10\% & 3.23\%  & 3.14\% & 3.87\% \\
Acceptance rate of Grow & 
1.50\%  & 1.36\%  & 0.90\%  & 0.65\%  \\
Acceptance rate of Prune & 1.43\%  & 1.20\% & 0.54\% & 0.30\%  \\
Acceptance rate of Change1 & -  & - & 6.09\%  & 8.19\%  \\
Acceptance rate of Change2 & 4.17\%  & 4.87\% & -  & 5.66\%  \\
\yz{Acceptance} rate of Swap & - & 4.01\%  & 3.49\% & 4.52\%  \\

  \bottomrule  

\end{tabular}

 \label{table_S1-1_IteTim}
\end{table}

\llj{We fix $\gamma=0.99$ and $\rho=15$, as for Figure \ref{Fig:S1_trace plots}. For each of the experiments E1--E4, we run the P-BCART MCMC algorithm 10 times and for each run we record the   
iteration time  
until an ``optimal'' tree is found. The average iteration time with the standard deviation (s.d.) of the 10 runs  and the average acceptance rates of 
moves are shown in Table \ref{table_S1-1_IteTim}. The figures in the second row indicate that the experiment E4 is faster in finding an ``optimal'' tree than E1--E3 when at least one of the Change moves or/and the Swap move is removed. In particular, the comparison between E1 and E2 confirms the essence of the Swap move, as illustrated also in \cite{chipman1998bayesian}. 
Moreover, 
the acceptance rate of all moves is a weighted average of  acceptance rates of all individual moves, and we observe that the acceptance rates of the Change and Swap moves (in particular, the Change1 move) are significantly greater than Grow and Prune moves, 
which also confirms the significance of the Change and Swap moves (especially, the Change1 move).} 

We also ran several other similar but more complex simulation examples to check the performance of P-BCART, NB-BCART and ZIP-BCART \yz{models}. Our conclusions from these simulations are: 1) BCART models can retrieve the tree structure (including both topology and \lj{parameters}) as that used to simulate the data, 2) BCART \yz{models are able to avoid choosing} noise variables \yao{regardless of their distributions, and 3) \llj{the Change and Swap moves have significant impacts on the BCART models and it is beneficial to include two types of the Change move}}.

\subsubsection{Scenario 2: ZIP data with varying probability of zero mass component}

\COM{
\textbf{Small probability of zero}
	\begin{itemize}
	\item $i=1,...,5000.$
    \item Simulate $x_{i1} \sim N(0,1)$.
    \item Simulate $x_{i2} \sim N(0,1)$.
    \item Simulate $N_{i}$ using $\lambda(x_{i1}, x_{i2})$ in Zero-Inflated Poisson distribution.
    $$
	\lambda \left(x_{i1},x_{i2}\right)=\left\{\begin{array}{ll}
	7 & \text { if } x_{i1}x_{i2} \leq 0 \\
	1 & \text { if } x_{i1}x_{i2} > 0
	\end{array}\right.
	$$
\item Probability of zero = 0.05. 
\end{itemize}
}

We simulate a data set $\{(\vk{x}_i,v_i,N_i)\}_{i=1}^n$ with $n=5,000$ independent observations. Here $\vk{x}_i=(x_{i1}, x_{i2})$, with $x_{ik} \sim N(0,1)$ for $k=1,2$. \ji{We assume $v_{i}\equiv 1$ for simplicity, since it is not a key feature in this Scenario.}
Moreover, $N_i\sim \text{ZIP}(p_0,\lambda \left(x_{i1},x_{i2}\right) )$, where
 $$
	\lambda \left(x_{1},x_{2}\right)=\left\{\begin{array}{ll}
	7 & \text { if } x_{1}x_{2} \leq 0, \\
	1 & \text { if } x_{1}x_{2} > 0,
	\end{array}\right.
	$$
and $p_0\in(0,1)$ is the probability of a zero due to the point mass component, for which the value is to be specified. 
The data is split into two subsets: a training set with $n-m=4,000$ observations and a test set with $m=1,000$ observations.

For this Scenario, we aim to examine how the P-BCART, NB-BCART and ZIP-BCART will perform when $p_0$ is varied. Note that since \yz{$v_i \equiv 1$}, NB1 and NB2 (ZIP1 and ZIP2) will be essentially the same. Intuition tells us that when $p_0$ is small NB-BCART should be \yz{good enough} to capture the over-dispersion introduced by a small proportion of zeros, but when $p_0$ becomes large ZIP-BCART should perform better for the highly over-dispersed data. This intuition will be confirmed by \lj{this study}. For simplicity, we shall present two results, one with $p_0=0.05$ and the other with 
$p_0=0.95$.

\begin{table}[!t] 

 \centering
\caption{Hyper-parameters, \yz{$p_D$ (or $q_D$,$r_D$) and DIC} \yz{on} training data ($p_0=0.05$). Bold font indicates DIC selected model.}

\begin{tabular}{lcccc} 

\toprule   

  Model & $\gamma$ & $\rho$ & $p_{D}$(or $q_D, r_D$) & \multicolumn{1}{c}{DIC} \\  

\midrule   

ZIP-BCART (2) & 0.50 &  20 & 4.00 &  
11451\\
ZIP-BCART (3) & 0.99 & 20 & 5.94 &  
11405\\
\textbf{ZIP-BCART (4)} & 0.99 & 15 & {7.95} & \textbf{
11322}  \\
ZIP-BCART (5) & 0.99 & 5 & 9.86 & 
11364 \\\hline

P-BCART (2) & 0.50 &  20 & 2.00 &  
11369\\
P-BCART (3) & 0.99 & 20 & 2.99 & 
11337\\
\textbf{P-BCART (4)} & 0.99 & 10 & {3.99}  & \textbf{
11262}  \\
P-BCART (5) & 0.99 & 5 & 4.91 &  
11299\\\hline

NB-BCART (2) & 0.50 & 30& 4.00 &  
11317\\
NB-BCART (3) & 0.99 & 25& 5.99 &  
11273\\
\textbf{NB-BCART (4)} & 0.99 & 20 & {7.99} &  \textbf{
11192}\\
NB-BCART (5) & 0.99 & 5 & 9.90 &  
11237\\

  \bottomrule  

\end{tabular}

 \label{table-S2-1}
\end{table}

\begin{table}[!t]  

 \centering
\caption{
\yz{
Model performance on test data
}($p_0=0.05$) with bold entries determined by DIC (see Table \ref{table-S2-1}).}
 \begin{tabular}{lccccc}

\toprule   

   Model & RSS($\vk{N}$) & RSS($\vk{N/v}$) & \ji{NLL} & DS($\vk{N/v}$) & \multicolumn{1}{c}{Lift}  \\  

\midrule   

ZIP-BCART (2) & 
2013 & 
0.00222 & 
1975 & 0.000185
 & 
1.22\\
ZIP-BCART (3) & 
1986 & 
0.00208 & 
1953 & 0.000169
 & 
2.67 \\
\textbf{ZIP-BCART (4)} & 
1923 & \textbf{
0.00162} & 
1890 & \textbf{0.000116
} & 
6.34 \\
ZIP-BCART (5) & 
1909 & 
0.00182 & 
1863 & 0.000130
 & 
6.56 \\\hline

P-BCART (2) & 
1758 & 
0.00175 & 
1702 & 0.000138
& 
1.40 \\
P-BCART (3) & 
1732 & 
0.00160 & 
1673 & 0.000123
 & 
3.21 \\
\textbf{P-BCART (4)} & 
1681 & \textbf{
0.00108} & 
1612 & \textbf{0.000072} & 
6.62\\
P-BCART (5) & 
1662 &  
0.00126 & 
1594 & 0.000092 & 
6.75 \\\hline

NB-BCART (2) & 
1683 & 
0.00145 & 
1647 & 0.000101 & 
1.58\\
NB-BCART (3) & 
1661 & 
0.00131 & 
1616 & 0.000092 & 
3.53 \\
\textbf{NB-BCART (4)} & 
1609 & \textbf{
0.00070} & 
1536 &  \textbf{0.000056} & 
6.95\\
NB-BCART (5) & 
1589 & 
0.00097 & 
1502 &  0.000074 & 
6.97\\

  \bottomrule  

\end{tabular}

 \label{table-S2-2}
\end{table}


We first discuss the simulation with a small probability of zero  mass (i.e., $p_0=0.05$). In Table \ref{table-S2-1} we present the hyper-parameters $\gamma, \rho$ used to obtain MCMC convergence to the region of trees with a certain number of terminal nodes (indicated after the abbreviation of models, e.g., the 2 in ZIP-BCART (2)).  \yz{The last two columns give the effective number of parameters and DIC of the optimal trees for each model, respectively.}
We can conclude from the DIC that by using Step 3 in Table \ref{table_SS}  we can select the optimal tree with  
the true 4 terminal nodes for either ZIP-BCART, P-BCART or NB-BCART, and among those, the NB-BCART (with DIC=11192) is the best one. This looks a bit surprising at a first glance because our data are simulated from a ZIP model. We suspect that the reason for this may be two-fold: First, the NB is enough to capture the small over-dispersion. 
Second,  we have used data-augmentation in the algorithms and thus it is understandable that the NB-BCART with 1 latent variable (\yao{see Section \ref{NB}}) could achieve better performance than the ``real" ZIP-BCART with 2 latent variables (\yao{ see Section \ref{ZIP}}). Moreover, we see that even the P-BCART performs better than the ZIP-BCART, for similar reasons.

Now, let us look at  the performance of these models on test data in Table \ref{table-S2-2}. First, we see that for each type of model, ZIP, Poisson and NB, the optimal tree with 4 terminal nodes achieves best RSS($\vk{N/v}$) (0.00162, 0.00108 and 0.00070, respectively) and DS($\vk{N/v}$) (0.000116, 0.000072 and 0.000056, respectively) \yz{on} test data, which is not surprising as those models retrieve the almost true tree structures. 
Second, we see from  RSS($\vk{N}$) that for each type of model, the performance becomes better as the number of terminal nodes that we want increases, however, the amount of decrement becomes smaller after the optimal trees with 4 terminal nodes have been obtained. 
We  observe the same for \ji{negative log-likelihood and} lift.
It is worth noting that when calculating and comparing lift for different trees, instead of simply following the four \ji{steps} in {\bf M5}, in Step 4 we first choose the minimum total sum of exposures among the least and most risky groups in all the trees to be compared and then calculate other values accordingly using this minimum \ji{total sum of} exposures as the basis. 
Third, we see that among these three trees with 4 terminal nodes\yz{,} the one obtained from NB-BCART gives the best performance \yz{on} test data based on all these performance measures, which is consistent with the conclusion from training data.

Next, we consider the simulation with a large probability of zero  mass (i.e., $p_0=0.95$). The results are displayed in Tables \ref{table-S2-3} and  \ref{table-S2-4}. Similar \ji{discussions} can be done for this case. In particular, we find that the performance order based on DIC is ZIP-BCART$>$NB-BCART$>$P-BCART, which is also  \ji{consistent with their performance on test data}. 


We also ran several other  similar simulation examples to check the performance of P-BCART, NB-BCART and ZIP-BCART with different values for $p_0$. Our conclusion from these simulations is that when the proportion of zeros in the data is small (reflected by small $p_0$) then the NB-BCART or P-BCART performs better than ZIP-BCART, whereas when the proportion of zeros in the data is large then the ZIP-BCART is preferred to NB-BCART \yz{and P-BCART}. This finding is consistent with the real insurance data discussed below.

\begin{table}[!t] 

 \centering
\caption{Hyper-parameters, \yz{$p_D$ (or $q_D$,$r_D$) and DIC} \yz{on} training data ($p_0=0.95$). Bold font indicates DIC selected model.}
 \label{table-S2-3}

\begin{tabular}{lcccc} 

\toprule   

  Model & $\gamma$ & $\rho$ & $p_{D}$(or $q_D, r_D$) & \multicolumn{1}{c}{DIC} \\  

\midrule   

ZIP-BCART (2) & 0.50 &  10  & 3.99 &  
3483\\
ZIP-BCART (3) & 0.99 & 10& 5.99 & 
3452 \\
\textbf{ZIP-BCART (4)} & 0.99 & 8 & 7.95 & \textbf{
3375} \\
ZIP-BCART (5) & 0.99 & 3  & 9.93 &  
3396 \\\hline

P-BCART (2) & 0.50 &  10 & 1.98 & 
3892 \\
P-BCART (3) & 0.99 & 10 & 2.96 &  
3863\\
\textbf{P-BCART (4)} & 0.99 & 5 & 3.91 &   \textbf{
3801} \\
P-BCART (5) & 0.99 & 2 & 4.90 &   
3827\\\hline

NB-BCART (2) & 0.50 & 20 & 3.99 &  
3726\\
NB-BCART (3) & 0.99 & 20& 5.97 &  
3699 \\
\textbf{NB-BCART (4)} & 0.99 & 10 & 7.92 &  \textbf{
3632} \\
NB-BCART (5) & 0.99 & 8 & 9.89 &  
3667\\

  \bottomrule  

\end{tabular}

\end{table}

\begin{table}[!t]  

 \centering
\caption{
\yz{
Model performance on test data
} ($p_0=0.95$)  with bold entries determined by DIC (see Table \ref{table-S2-3}).}

 \begin{tabular}{lccccc} 

\toprule   

   Model & RSS($\vk{N}$) & RSS($\vk{N/v}$) & NLL & DS($\vk{N/v}$) & Lift \\  

\midrule   
ZIP-BCART (2) & 
721 & 
0.00755 & 
699 & 
0.00721 & 
1.25 \\
ZIP-BCART (3) & 
715 & 
0.00700 & 
690 &  
0.00698 & 
1.92 \\
\textbf{ZIP-BCART (4)} & 
682 & \textbf{
0.00571} & 
657 & \textbf{
0.00619} & 
2.86 \\
ZIP-BCART (5) & 
675 & 
0.00613 & 
649 &  
0.00646 & 
3.13 \\\hline
 
P-BCART (2) & 
782 & 
0.00967 & 
754 & 
0.00802 & 
1.15 \\
P-BCART (3) & 
773 & 
0.00891 & 
746 & 
0.00786 & 
1.50 \\
\textbf{P-BCART (4)} & 
750 & \textbf{
0.00723} & 
719 & \textbf{
0.00712} & 
2.40 \\
P-BCART (5) & 
741 &  
0.00792 & 
705 & 
0.00739 & 
2.72 \\\hline

NB-BCART (2) & 
775 & 
0.00893 & 
740 & 
0.00775 & 
1.19 \\
NB-BCART (3) & 
768 & 
0.00810 & 
731 & 
0.00740 & 
1.72 \\
\textbf{NB-BCART (4)} & 
735 & \textbf{
0.00647} & 
701 &  \textbf{
0.00667} & 
2.60\\
NB-BCART (5) & 
730 & 
0.00703 & 
693 &  
0.00689 & 
2.90 \\

  \bottomrule  

\end{tabular}

 \label{table-S2-4}
\end{table}

\subsubsection{Scenario 3: Different ways to incorporate exposure in ZIP models}


The purpose of Scenario 3 is to compare two different ways of dealing with exposure, namely, ZIP1-BCART and ZIP2-BCART.
To this end, we simulate a data set  $\{(\vk{x}_i,v_i,N_i)\}_{i=1}^n$ with $n=5,000$ independent observations. Here $v_{i}\sim \text{U}(0,1)$, $\vk{x}_i=(x_{i1}, x_{i2})$, with $x_{ik} \sim N(0,1)$ for $k=1,2$. Moreover, $N_i\sim \text{ZIP}(p_{i}^{(\tau)},\lambda \left(x_{i1},x_{i2}\right) v_i)$, where
 $$
	\lambda \left(x_{1},x_{2}\right)=\left\{\begin{array}{ll}
	7 & \text { if } x_{1}x_{2} \leq 0, \\
	1 & \text { if } x_{1}x_{2} > 0,
	\end{array}\right.
	$$
and the probability of zero mass component is given as
$$p^{(\tau)}_i =\frac{\mu(x_{i1}, x_{i2})}{v_{i}^\tau +\mu(x_{i1}, x_{i2})}, \ \ \text{with} \ \ \mu(x_{i1}, x_{i2})\equiv 0.5,$$
and some $\tau\ge0$ to be specified below.
The data is split into two subsets, namely a training set with $n-m=4,000$ observations and a test set with $m=1,000$ observations.

In the above simulation setup, we include exposure in both the Poisson  component and the zero mass component. In this way, it is not clear which of ZIP1-BCART and ZIP2-BCART will outperform the other. That being said, we could vary the value of $\tau$ to control the effect of exposure to the zero  mass component. We shall consider two extreme cases, one with a very small $\tau$ and the other with a very large $\tau$. More precisely, for a large $\tau$  we choose $\tau=100$. In this case, since \ji{many} $v_{i}^\tau$ will be  small, we have that $p^{(\tau)}_i$ will be close to one, which implies that Poisson component should play a minor role \ji{in exposure modelling} and thus we would expect that ZIP2-BCART has better ability to capture this. On the other hand, for a small value $\tau=0.0001$, since \ji{many} $v_{i}^\tau$ will be close to 1 we have that $p^{(\tau)}_i$ will be \ji{almost} independent of $v_i$, which implies that zero mass component should play a minor role \ji{in exposure modelling} and thus we would expect that ZIP1-BCART has better ability to capture this.  We report DIC for these two cases in Table \ref{table-S3-1}. The model performances on test data are listed in Table \ref{table-S3-2} for $\tau=100$ and Table \ref{table-S3-3} for $\tau=0.0001$. \ji{From these tables, we can confirm the above intuition that ZIP1-BCART should perform better for small $\tau$ and worse for large $\tau$ (compared to ZIP2-BCART). We conclude from this simulation study  that the ZIP2-BCART works better in capturing the potential stronger effect of the exposure to the zero mass component, which is also illustrated in the real insurance data discussed below.}



\begin{table}[!t] 

 \centering
\caption{DIC for ZIP-BCART with different values of $\tau$ \yz{on training data}. Bold font indicates DIC selected model.}

\begin{tabular}{lrr} 

\toprule   

  Model  & DIC ($\tau=100$) & DIC ($\tau=0.0001$)   \\  

\midrule   

ZIP1-BCART (2)  &  
3091 & 
10515 \\
ZIP1-BCART (3) &  
3055 &  
10437 \\
\textbf{ZIP1-BCART (4)}  & \textbf{
2976} & \textbf{
10273}\\
ZIP1-BCART (5)  &  
2997 & 
10330 \\\hline

ZIP2-BCART (2)  &  
2653 & 
10924 \\
ZIP2-BCART (3)  &  
2637 & 
10843 \\
\textbf{ZIP2-BCART (4)}  & \textbf{
2613}& \textbf{
10685} \\
ZIP2-BCART (5)  &   
2627 &  
10751
 \\ 

  \bottomrule  

\end{tabular}

 \label{table-S3-1}
\end{table}

\begin{table}[!t]  

 \centering
\caption{Model performance on test data ($\tau=100$) with bold entries determined by DIC (see Table \ref{table-S3-1}).}

 \begin{tabular}{lccccc} 

\toprule   

   Model & RSS($\vk{N}$) & RSS($\vk{N/v}$) (in 10$^{-5}$) & NLL & DS($\vk{N/v}$) & \multicolumn{1}{c}{Lift}  \\  

\midrule   

ZIP1-BCART (2) & 
2423 & 
3.06 & 
1339 & 
0.00281 & 
1.01\\

ZIP1-BCART (3) & 
2417 & 
2.98 & 
1330 & 
0.00259 & 
1.33 \\

\textbf{ZIP1-BCART (4)} & 
2376 & \textbf{
2.20} & 
1308 & \textbf{
0.00209}   & 
1.78\\

ZIP1-BCART (5) & 
2333 & 
2.63 & 
1302 &   
0.00219 & 
1.81 \\\hline

ZIP2-BCART (2) & 
2072 & 
2.76& 
1324 & 
0.00234 & 
1.06\\

ZIP2-BCART (3) & 
2069 & 
2.57& 
1317 & 
0.00207 & 
1.46\\

\textbf{ZIP2-BCART (4)} & 
2056 & \textbf{
2.02} & 
1304 & \textbf{
0.00179} & 
1.97\\

ZIP2-BCART (5) & 
2049 & 
2.06& 
1295 &  
0.00189 & 
2.08\\

  \bottomrule  

\end{tabular}

 \label{table-S3-2}
\end{table}

\begin{table}[!t]  

 \centering
\caption{Model performance on test data ($\tau=0.0001$) with bold entries determined by DIC (see Table \ref{table-S3-1}).}

 \begin{tabular}{lccccc} 

\toprule   

 Model & RSS($\vk{N}$) & RSS($\vk{N/v}$) & \ji{NLL} & DS($\vk{N/v}$) & \multicolumn{1}{c}{Lift} \\  

\midrule

ZIP2-BCART (2) & 
6859 & 
0.0093 & 
4185 & 
0.0080 & 
1.02\\

ZIP2-BCART (3) & 
6648 & 
0.0080 & 
4092 & 
0.0069 & 
2.10\\

\textbf{ZIP2-BCART (4)} & 
6408 & \textbf{
0.0060} & 
3913 & \textbf{
0.0050} & 
3.40\\

ZIP2-BCART (5) & 
6320 & 
0.0073 & 
3853 & 
0.0062 & 
3.48 \\\hline

ZIP1-BCART (2) & 
6628 & 
0.0079 & 
3827 & 
0.0072 & 
1.07\\

ZIP1-BCART (3) & 
6535 &  
0.0058 & 
3763 & 
0.0055 & 
2.15\\

\textbf{ZIP1-BCART (4)} & 
6350 & \textbf{
0.0027} & 
3590 & \textbf{
0.0024}  & 
3.45\\

ZIP1-BCART (5) & 
6282 & 
0.0036 & 
3543 & 
0.0033 & 
3.62\\

  \bottomrule  

\end{tabular}

 \label{table-S3-3}
\end{table}


\subsection{Real data analysis}
We illustrate our methodology with a real insurance dataset, named \textit{dataCar}, available  from the library \texttt{insuranceData} in \textsf{R}; see \cite{R:insurance} for details.  This dataset is based on one-year vehicle insurance policies taken out in 2004 or 2005. There are 67,856 policies of which 93.19\% made no claims. A summary of the variables used is given in Table \ref{table_cardata}. We split this dataset into training (80\%) and test (20\%) data sets, in doing so we keep the balance of zero and non-zero claims in both training and test data sets.

\begin{table}[ht]

 \centering
 
\caption{Description of variables (\textit{dataCar})}

\begin{tabular}{p{2cm}p{9cm}p{2cm}} 

\toprule   

  Variable  & Description & Type \\  

\midrule   
 numclaims  & number of claims & numeric\\
 exposure  &in yearly units, between 0  and 1 & numeric\\
   veh\_value  &vehicle value, in \$10,000s & numeric\\  

  veh\_age   & vehicle age category, 1 (youngest), 2, 3, 4 & numeric \\
   
agecat & driver age category, 1 (youngest), 2, 3, 4, 5, 6 & numeric\\

veh\_body  &vehicle body, include 13 different types coded as 
\ji{HBACK, UTE, STNWG, HDTOP, PANVN, 
SEDAN, TRUCK, COUPE, MIBUS,  MCARA, 
BUS, CONVT, RDSTR} &character\\

gender  & Female or Male &character\\

area  & coded as A B C D E F  &character\\

  \bottomrule  

\end{tabular}
\label{table_cardata}

\end{table}

\begin{table}[!t]  

 \centering
\caption{Hyper-parameters, $p_D$ (or $q_D$,$r_D$) and DIC on training data (\textit{dataCar}). Bold font indicates DIC selected model.}

 \begin{tabular}{lrrrr}

\toprule   

   Model & \multicolumn{1}{c}{$\gamma$} & $\rho$ & \multicolumn{1}{c}{$p_{D}$(or $q_D$,$r_D$)} & \multicolumn{1}{c}{DIC} \\  

\midrule

P-BCART (4) & 0.99 & 15 & 4.00 & 27948.8  \\

\textbf{P-BCART (5)} & 0.99 & 8 & {5.00} & \textbf{27943.8}  \\

P-BCART (6) & 0.99 & 6 & 6.00 & 27944.4 \\\hline

NB1-BCART (4) & 0.99 & 15 & 7.98 & 26002.4  \\

\textbf{NB1-BCART (5)} & 0.99 & 7 & {9.96} & \textbf{25892.0}   \\

NB1-BCART (6) & 0.99 & 6 &   11.96 & 25945.2\\\hline

NB2-BCART (4) & 0.99 & 15 & 7.99 & 25925.7   \\

\textbf{NB2-BCART (5)} & 0.99 & 6 & {9.98} & \textbf{25846.4}  \\

NB2-BCART (6) & 0.99 & 5 &  11.97 &   25885.6 \\\hline

ZIP1-BCART (4) & 0.99 & 10 & 8.05 & 25688.4  \\

\textbf{ZIP1-BCART (5)} & 0.99 & 5 & {9.85} & \textbf{25674.1}  \\

ZIP1-BCART (6) & 0.99 & 3 & 12.00 & 25678.3  \\\hline

ZIP2-BCART (4) & 0.99 & 10 & 7.99 & 25654.3  \\

\textbf{ZIP2-BCART (5)} & 0.99 & 4 & {9.91} & \textbf{25632.5}   \\

ZIP2-BCART (6) & 0.99 & 3 & 11.93 & 25641.4   \\

  \bottomrule  

\end{tabular}

 \label{table_Car-T}
\end{table}

We shall apply the BCART \ji{models} for claims frequency \yz{modelling} introduced in Section \ref{Sec:CF} to training data, where we can use the three-step approach given in Table \ref{table_SS} to choose an optimal tree for each model (and also a global optimal one). We then assess the performance of these obtained trees  on test data.

\begin{figure}[!htb]
    \centering
    \begin{minipage}{.5\textwidth}
        \centering
        \includegraphics[width=0.9\textwidth]
        {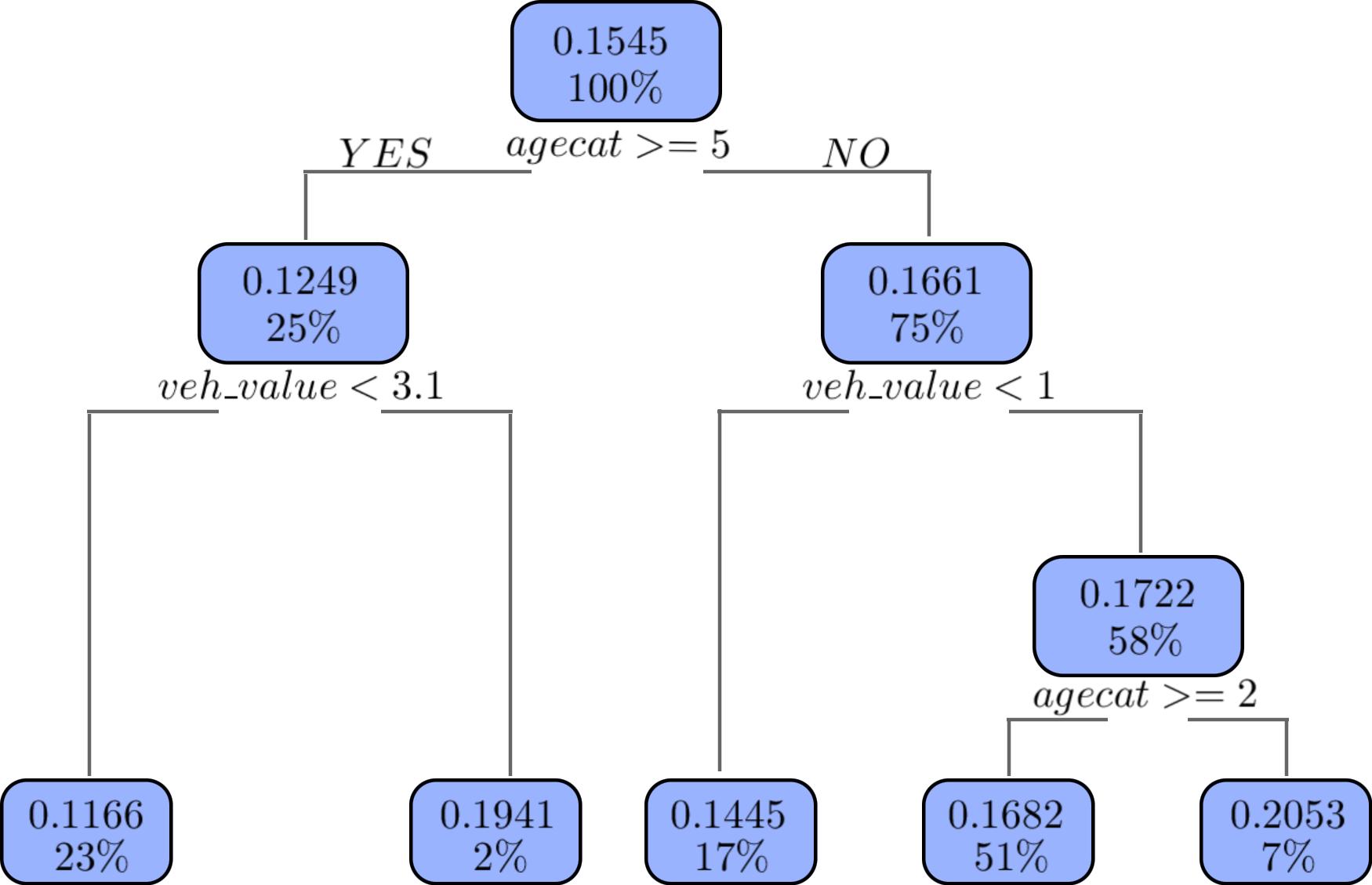}
        \caption{Tree from P-CART. Numbers at \\ each node give the estimated frequency and \\ the percentage of observations.}
        \label{Fig:real data cart}
    \end{minipage}%
    \begin{minipage}{0.5\textwidth}
        \centering
        \includegraphics[width=0.9\linewidth]{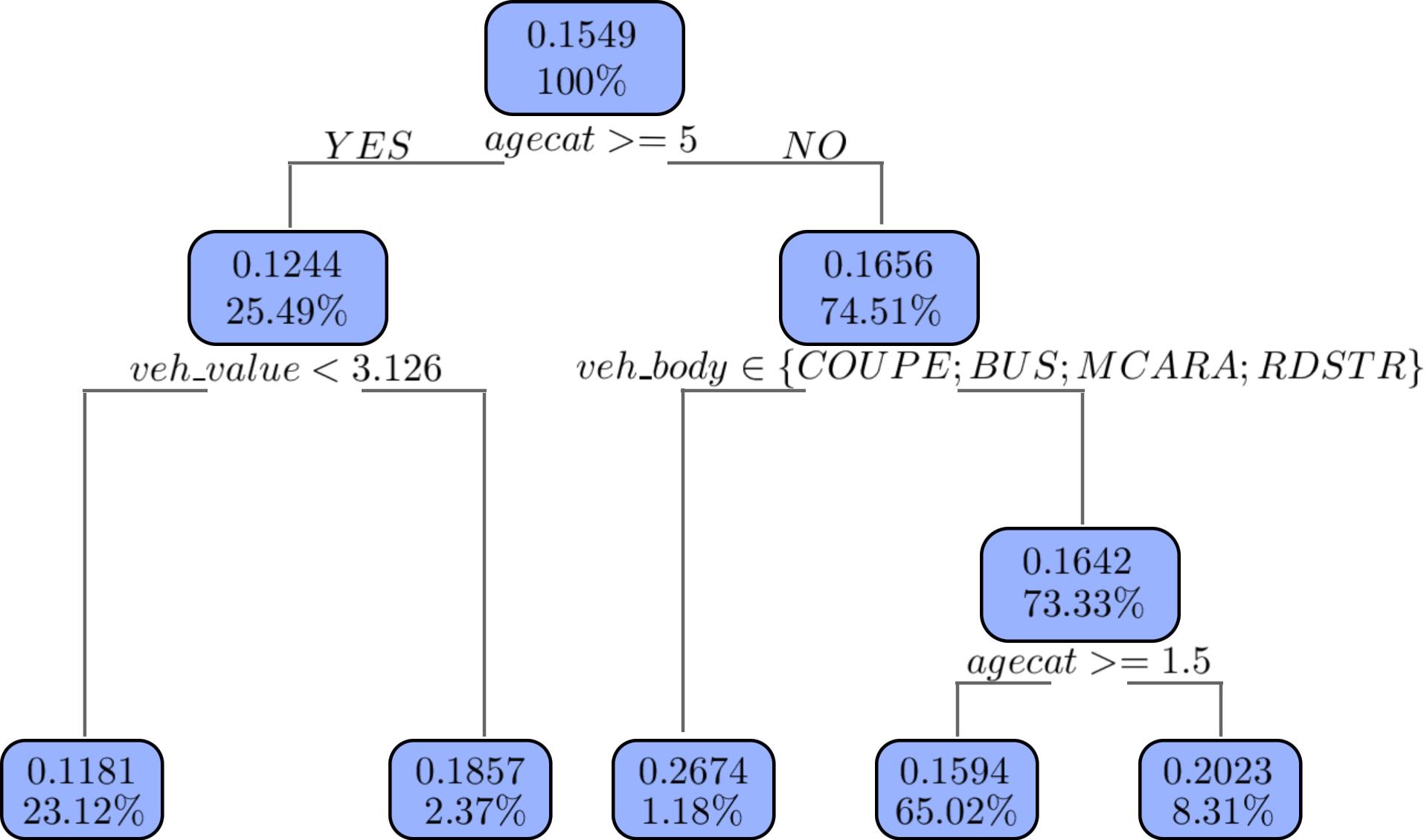}
        \caption{Optimal tree from ZIP2-BCART. Numbers at each node give the estimated frequency and the percentage of observations.}
        \label{Fig:real data ZIP2-BCART}
    \end{minipage}
\end{figure}

\yao{Running ANOVA-CART on the training data, we use cross-validation to select the tree size, which has 5 terminal nodes. \yao{We also run P-CART in the same way, again resulting in a tree with 5 terminal nodes, and this tree is} shown in Figure \ref{Fig:real data cart}.} 
Then, we apply P-BCART, NB1-BCART, NB2-BCART, ZIP1-BCART and ZIP2-BCART to  \yz{the} same data. Based on the knowledge learnt from \yz{CARTs} above, we can tune the hyper-parameters $\gamma, \rho$, so that the algorithm will converge to a region of trees with number of terminal nodes around 5. Some of these, together with the effective number of parameters and  DIC, are shown in Table \ref{table_Car-T}. We see from this table that all the effective numbers of parameters are reasonable for the model used to fit the data.  
We conclude from the DIC that all of these BCART models  
select an optimal tree with 5 terminal nodes using the three-step approach, and among these the one from ZIP2-BCART, with the smallest 
 DIC(=25632.5), should be chosen as the global optimal tree to characterize the data.

It is interesting to check whether there are similarities in the trees obtained from different models, including the P-CART, particularly as they all have 5 terminal nodes. \ji{For the tree from P-CART illustrated in Figure \ref{Fig:real data cart},} the variable“\textit{agecat}” is first used and then “\textit{veh\_value}”, followed by “\textit{agecat}”  again. The tree from P-BCART \ji{(not shown here)} also uses “\textit{agecat}” first, 
but in the following steps, it uses “\textit{veh\_value}” and “\textit{veh\_body}”. The trees from NB1-BCART and NB2-BCART look very similar,  \yz{and} both of them use “\textit{gender}” first and then  use “\textit{agecat}”, “\textit{veh\_value}” and “\textit{veh\_body}”. 
Further, the trees from  ZIP1-BCART and ZIP2-BCART  have the same tree structure and \yz{select} \ji{the same} \yz{splitting} variables as the tree from P-BCART, while the split \yz{values/categories} \yz{are slightly different}. The optimal tree from ZIP2-BCART is displayed in Figure \ref{Fig:real data ZIP2-BCART}, where the estimated frequency (i.e., the first figure in each node) is calculated through \eqref{eq:yhat} for the ZIP2 model  with unit exposure. Comparing the two trees  in Figures \ref{Fig:real data cart} and \ref{Fig:real data ZIP2-BCART} we see that ZIP2-BCART model can identify a more risky group (i.e., the one with estimated frequency equal to 0.2674).
 Moreover, \ji{for comparison}
we also use GLM to fit the data. We find that only the variables “\textit{agecat}” and “\textit{veh\_body}” are significant, in which we also use the interactions between these two variables. In conclusion, though the variables used for different models \ga{can differ slightly, there seems to be a consensus that} “\textit{agecat}”, “\textit{veh\_value}” and “\textit{veh\_body}” are relatively significant variables and “\textit{gender}”, “\textit{veh\_age}” and “\textit{area}” are less significant.  

\begin{table}[!t]  

 \centering
\caption{\yao{Model performance on test data (\textit{dataCar}) with bold entries determined by DIC (see Table \ref{table_Car-T}).}}

 \begin{tabular}{lp{1.3cm}p{1.1cm}p{1.2cm}p{1.2cm}p{1cm}p{1.3cm}p{1.3cm}} 

\toprule   

   Model & RSS ($\vk{N}$) & RSS ($\vk{N/v}$) & \multicolumn{1}{c}{\ji{NLL}} & DS($\vk{N/v}$) & \multicolumn{1}{c}{Lift} & 
   \yao{Time (s)} & \yao{Memory 
   (MB)}  \\  

\midrule   

GLM & 1057.029 & - & 
5532.37 & - & - & \phantom{62}1.15 & 115  \\\hline

\yao{ANOVA-CART (5)} & 1054.061 & 
0.0205 & 
5514.06 & 
0.0700 & 
1.83 & \phantom{62}2.05 & 
\phantom{3}98\\\hline

P-CART (5) & 1042.295 & 
0.0185 & 
5476.43 & 
0.0681 & 
1.97 & \phantom{32}2.13 & 
\phantom{3}98\\\hline

P-BCART (4) & 1042.221 & 
0.0172 & 
5473.90 & 
0.0680 & 
1.74 & 317.61 & 364  \\
\textbf{P-BCART (5)} & 1042.211 & \textbf{
0.0167} & 
5472.86 & \textbf{
0.0602} & 
2.26 & 291.28 & 378 \\
P-BCART (6) & 1042.205 & 
0.0171 & 
5472.27 & 
0.0632 &  
2.29 & 325.10 & 581 \\\hline

NB1-BCART (4) & 1041.129 & 
0.0168 & 
5470.12 &  
0.0445 & 
1.80 & 413.95 & 628  \\
\textbf{NB1-BCART (5)} & 1041.109 & \textbf{
0.0159} & 
5469.00 &  \textbf{
0.0372}  & 
2.46 & 403.84 & 569 \\
NB1-BCART (6) & 1041.103 & 
0.0162 & 
5468.51 &  
0.0413 & 
2.57 & 459.70 & 689 \\\hline

NB2-BCART (4) & 1041.127 & 
0.0155 & 
5470.01 &  
0.0416 & 
1.85 & 431.90 & 642   \\
\textbf{NB2-BCART (5)} & 1041.102 & \textbf{
0.0144} & 
5468.68 &  \textbf{
0.0352}  & 
2.50 & 441.82 & 661  \\
NB2-BCART (6) & 1041.094 & 
0.0151 & 
5468.35 &  
0.0390 & 
2.58 & 492.19 & 721  \\\hline

ZIP1-BCART (4) & 1041.102 & 
0.0150 & 
5469.07 &  
0.0383 & 
1.91 & 548.29 & 827  \\
\textbf{ZIP1-BCART (5)} & 1041.087 & \textbf{
0.0138} & 
5468.39 &  \textbf{
0.0316}  & 
2.56 & 524.84 & 792  \\
ZIP1-BCART (6) & 1041.075 & 
0.0142 & 
5468.02 &  
0.0362 & 
2.60 & 569.21 & 889  \\\hline

ZIP2-BCART (4) & 1041.054 & 
0.0145 & 
5468.25 &  
0.0279 & 
2.20 & 561.98 & 840  \\
\textbf{ZIP2-BCART (5)} & 1041.038 & \textbf{
0.0136} & 
5468.01 & \textbf{
0.0241}  & 
2.72 & 570.40 & 851 \\
ZIP2-BCART (6) & 1041.025 & 
0.0141 & 
5467.81 &  
0.0271 & 
2.79 & 589.24 & 892   \\

  \bottomrule  

\end{tabular}

 \label{table-Car-test}
\end{table}

Now, we apply the trees 
to the test data. The performances are given in Table \ref{table-Car-test}. We also include the commonly used GLM, for which the performance looks not as good as the tree models.
From the table, we can conclude that for each of the BCART models the tree with 5 terminal nodes that is selected by DIC performs better, in terms of RSS($\vk{N/v}$) and DS($\vk{N/v}$), than the trees with either smaller or larger number of terminal nodes. This confirms that the proposed three-step approach for \yz{the tree} model selection  in each type of models based on DIC works well in real data. Moreover, all the performance measures give the same ranking of models (from best to worst) as follows:
\centerline{\small \text{ZIP2-BCART}, \text{ZIP1-BCART}, \text{NB2-BCART}, \text{NB1-BCART}, \text{P-BCART}, \text{P-CART}, \text{ANOVA-CART}, \text{GLM.}}
This ranking is, to some extent, consistent with the conclusions from the simulation examples and as expected. We do not know the exact distribution of real insurance data, but we do know that it contains a  high proportion of zeros, where the advantage of ZIP 
comes into play. Further, comparing \ji{NB} and Poisson distributions, the former is able to handle over-dispersion, so their performance ranking is reasonable. Moreover, the ranking of two ZIP-BCART models and two NB-BCART models are also consistent with the conclusions of \cite{lee2020delta, lee2021addressing} where it is \ji{justified} that the non-standard ways of dealing with exposures (i.e., ZIP2-BCART and NB2-BCART) should better fit real insurance data.


In addition to \llj{the performance measures, we also record the 
 computation time (in seconds) and memory usage (in megabytes); see  the last two columns of Table \ref{table-Car-test}. All computations were performed on a laptop with Processor (3.5 GHz Dual-Core Intel Core i7) and Memory (16 GB 2133 MHz LPDDR3).
Clearly, BCART models are far inferior to CARTs and GLM in these two respects and as the number of latent variables increases (from P-BCART to NB-CART to ZIP-BCART) these indicators become worse, but we think with such a large training data these are still acceptable and feasible to use in practice. We remark that there have been
prior endeavors to address computing issues; see, e.g., \cite{chipman2014bayesian, he2019xbart,sparapani2021nonparametric}. We believe these two indicators will be improved after our code is optimized in the future. 
}

We conclude this section with some discussions on the {\it stability} of the proposed BCART models. Stability is a notion in computational learning theory of how the output of a machine learning algorithm is perturbed by small changes to its inputs. A stable learning algorithm is one for which the {prediction} does not change much when training data is modified slightly; see, e.g., \cite{APK19Stability} and references therein. CART models are known to be unstable. 
It is thus interesting to examine whether the proposed BCART models can be more stable. To this end, we propose the following approach to assess the stability of the P-CART and ZIP2-BCART (as the best) models. 
\begin{itemize}
    \item Randomly divide the data into two parts, 80\% for training and 20\% for \ji{testing}.
    \item Randomly select 90\% of training data for 20 times to construct 20 training subsets, named Data$_{1}$, Data$_{2}$, \ldots, Data$_{20}$.
    \item Obtain the optimal tree from P-CART  and ZIP2-CART, respectively, for each training set Data$_{j}$, $j=1,\ldots,20$. 
    \item Use the previously obtained trees to get predictions for test data. For each observation in test data, we will have 20 predictions from the 20 P-CART trees for which we calculate the variance, and do the same for the 20  ZIP2-BCART trees to get a variance.
    \item Calculate the mean (over the observations in test data) of those variances for P-CART and ZIP2-BCART, respectively.
    \end{itemize}
Since variance can capture the amount of variability, we shall use the above obtained mean to assess the stability (in their predicting ability) of a tree-based model. Namely, the smaller the mean the more stable the model that was used to calculate it. We apply it to the {\it dataCar} insurance data, the calculated mean for P-CART is 9.319339$\times 10^{-5}$ and  for ZIP2-BCART is 6.896231$\times 10^{-5}$. This implies that ZIP2-BCART is more stable than P-CART. Additionally, we also compare the 20 trees from P-CART, where we can observe very different trees in terms of number of terminal nodes (ranging from 3 to 8) and \yz{splitting} variables selected in the trees. Whereas, the 20 trees from ZIP2-BCART also show some stability in terms of number of terminal nodes (all around 5) and \yz{splitting} variables selected. 
The same procedure has also been applied to other BCART models and the conclusions are almost the same. Therefore, we conclude from our studies that the proposed BCART models in this paper show some stability that the CART models may not possess.

\section{Summary and discussions}

This work proposes the use of  BCART models for insurance pricing, and in particular, claims frequency prediction. These tree-based models  can automatically perform variable selection and detect non-linear effects and possible interactions among explanatory variables. The obtained optimal trees are relatively accurate, stable and are straightforward to interpret
by a visualization of the tree structure. These are desirable aspects for insurance pricing. 
We have introduced the  framework of the BCART model\yz{s} and presented  \yz{MCMC algorithms} for general non-Gaussian distributed data where data augmentation may be needed in its implementation. We have included BCART models for Poisson, \ji{NB} and ZIP distributions, which are the commonly used distributions for claims frequency. For the \ji{NB} and ZIP models, we explored two different ways to deal with exposures. 
Remarkably, we conclude from the simulation examples and real data analysis that the non-standard ways of embedding exposures can provide us with better tree models, which is in line with the conclusions of \cite{lee2020delta, lee2021addressing}. 
Furthermore, we introduced  a  tree model selection approach based on DIC, which has been seen to be an effective approach using both simulation examples and real insurance data. In particular,  we conclude from the real insurance data analysis that the ZIP-BCART with exposure embedded in the zero mass component is the best candidate for claims frequency \yz{modelling}. It is worth remarking that 
a general zero-inflated \ji{NB} BCART can be implemented and may further improve the accuracy, but this will require more latent variables to be introduced and will make the convergence of MCMC algorithm harder/slower; see \cite{murray2021log} for some insights. 

Below we 
comment on potential further improvements of the BCART models for claims frequency \yz{modelling}.
\begin{itemize}
    \item In the MCMC algorithms we have only used four common proposals, namely, Grow, Prune, Change, and Swap, which have made the algorithm to quickly converge to a local optimal region. In order to make it better explore the tree space, other proposals such as those in \cite{wu2007bayesian, pratola2016efficient} can be suggested to improve the mixing of simulated trees. However, we suspect this will significantly increase the computational time, particularly, for high-dimensional large data set and for models requiring data augmentation. To mitigate this effect, we might consider to use a non-uniform choice of splitting variables in the tree prior so as to achieve a better variable selection, e.g., the Dirichlet prior proposed in \cite{linero2018bayesian}.

    \item The proposed models have imposed several assumptions in order to simplify calculations. For example, we used conjugate prior for the terminal node distributions, and additional independence assumption as in \eqref{eq:int_lik_1} and \eqref{eq:ZIP1_integrated}. To further improve the
analysis, it might be beneficial to incorporate different specifications
of the prior for the same distribution scenario without using conjugate priors or independence, while this may require other techniques such as Laplace
approximation (see \cite{chipman2003bayesian}). We refer also to \cite{chipman2000hierarchical} for an  interesting incorporation of some hierarchical priors.
    \item We have proposed to use a single (optimal) tree induced from the BCART models for claims frequency prediction. The main reason for this choice is, as we discussed in the introduction, for ease of interpretation. Since stakeholders and regulators may not be statisticians \yz{who are} able to understand very complex statistical models, a single tree offers  intuitive and visual results to them. 
    Although we have proposed an approach to find one single optimal tree, some sub-optimal trees (in the convergence region of the MCMC) which possess similar/different tree structures, may also be as informative as the single optimal tree and should not be simply ignored. Further research can be done in this direction to make better use of the posterior trees by clustering or merging them; see, e.g.,  \cite{chipman2001managing, banerjee2012identifying}. 
    

    \item To further improve the accuracy of these Bayesian tree-based models we could explore BART for claims frequency \yz{modelling}. The BART models are tree ensemble\yz{s}; each tree in BART only accounts for a small part of the overall fit, potentially improving the performance, but  model interpretability needs to be explored before it can be used for insurance pricing. To this end, we believe  some insights  from  \cite{henckaerts2021boosting} would be helpful. 
    
\end{itemize}


In this paper, we have focused on insurance claims frequency. A natural next step is to construct a full insurance pricing BCART model, including both claims frequency and severity.

\bigskip

\llj{
{\bf Acknowledgement}:
We are  thankful to the anonymous referee for their constructive suggestions which have led to a significant improvement of the manuscript.
}

\printbibliography

\end{document}